%% file: 0.__Main.tex
\pdfoutput=1

\documentclass[11pt]{article}

\usepackage[final]{acl}

\usepackage{times}
\usepackage{latexsym}
\usepackage[T1]{fontenc}
\usepackage[utf8]{inputenc}
\usepackage{microtype}
\usepackage{multirow}
\usepackage{booktabs}
\usepackage{array}
\usepackage{inconsolata}
\usepackage{graphicx}
\usepackage{amsmath}
\usepackage{amssymb}
\usepackage{xcolor}
\usepackage{comment}
\usepackage{xspace}
\usepackage{enumitem}
\usepackage{adjustbox}
\usepackage{tcolorbox}
\usepackage{bbm}
\usepackage{dsfont}
\usepackage{hyperref}
\usepackage{pdfpages}

\newcommand{\chgd}[1]{\textcolor{black}{#1}}

\newcounter{promptcounter}
\renewcommand{\thepromptcounter}{\arabic{promptcounter}}
\newtcolorbox{MyBox}[2][]{%
  enhanced,
  breakable,
  colback=gray!5,
  colframe=gray!80!black,
  boxrule=1pt,
  toprule=2pt,
  rounded corners,
  arc=2pt,
  top=1.7mm,
  bottom=1.7mm,
  left=3mm,
  right=3mm,
  fuzzy shadow={0pt}{-2pt}{-0.5pt}{0.5pt}{black!35},
  title={\normalsize Prompt~\thepromptcounter.~#2}, 
  #1 
}

\title{Multi-level Diagnosis and Evaluation for Robust Tabular Feature Engineering with Large Language Models}

\author{Yebin Lim \\
  Computer Science and Engineering \\
  Korea University \\
  \texttt{yebinuni@korea.ac.kr} \\\And
  Susik Yoon \\
  Computer Science and Engineering \\
  Korea University \\
  \texttt{susik@korea.ac.kr} \\}

\begin{document}

\clearpage

\maketitle
\begin{abstract}
Recent advancements in large language models (LLMs) have shown promise in feature engineering for tabular data, but concerns about their reliability persist, especially due to variability in generated outputs. We introduce a multi-level diagnosis and evaluation framework to assess the robustness of LLMs in feature engineering across diverse domains, focusing on the three main factors: key variables, relationships, and decision boundary values for predicting target classes. We demonstrate that the robustness of LLMs varies significantly over different datasets, and that high-quality LLM-generated features can improve few-shot prediction performance by up to 10.52\%. This work opens a new direction for assessing and enhancing the reliability of LLM-driven feature engineering in various domains. Our source code is available at https://github.com/DohaLim/Robustness-eval.
\end{abstract}

\input{1.__Introduction}
\input{2.__Related_Work}

\input{3.__Methodology}

\input{4.__Experiments}

\input{5.__Conclusion}

\section*{Acknowledgments}
This work was partly supported by Korea University - KT (Korea Telecom) R\&D Center, the Institute of Information \& Communications Technology Planning \& Evaluation (IITP)-ICT Creative Consilience Program (IITP-2025-RS-2020-II201819), IITP-ITRC (Information Technology Research Center) (IITP-2025-RS-2024-00436857), Artificial Intelligence Star Fellowship Program (IITP-2025-RS-2025-02304828), and the National Research Foundation of Korea (NRF) (RS-2024-00406320) funded by the Korea government (MSIT).

\section*{Limitations}
\label{sec:limitations}
Despite the promising findings of this work, we acknowledge several limitations that can guide future research. First, the robustness of LLMs in feature engineering remains highly dependent on the characteristics of the underlying dataset. The variability observed across different domains suggests that LLMs may struggle with datasets that deviate significantly from their pre-trained knowledge. 

Additionally, while our multi-level evaluation framework with various sampling and corruption strategies provides insights into model-dataset reliability, it does not fully mitigate the risks associated with variation in quality and the selection of few-shot samples for inference. LLMs are still susceptible to generating features with incorrect relationships or suboptimal decision boundaries, depending on the given samples, which can negatively impact prediction performance.

Second, our study primarily evaluates LLM-driven feature engineering using binary features adopted by the state-of-the-art LLM-driven feature engineering method. While this approach is simple and effective, real-world applications often require more complex and intricate feature representations, where LLMs may exhibit even greater instability. Future research should explore strategies to address these challenges, such as focusing on zero-shot feature engineering or incorporating generated features of varying forms, allowing the proposed multi-level scheme to be further generalized and expanded.

Lastly, although our multi-level scheme is specifically designed to reveal the general-purpose model’s ability to handle variations in input conditions across diverse datasets, we intentionally kept the framework domain-agnostic to maximize its applicability. Nevertheless, we recognize that evaluating domain-specific LLMs could provide valuable insights into whether domain internalization enhances robustness in feature engineering. Such evaluations could further inform best practices and strengthen the reliability of LLM-driven methodologies across specialized application areas.


\newpage
\bibliography{References}

\newpage
\appendix
\input{6.__Appendix}

\end{document}

%% file: 1.__Introduction.tex
\section{Introduction}
Recent breakthroughs in large language models (LLMs) have opened new possibilities in tabular learning, such as feature engineering, question answering, and table comprehension~\cite{Fang}. 
The extensive pretrained knowledge of LLMs, when equipped with only a few examples, can automate costly data science workflows manually handled by domain experts. Notably, recent studies have shown that LLM-driven feature engineering can help outperform traditional tabular prediction methods, especially in a few- or zero-shot settings~\cite{FeatLLM, TabLLM, CAAFE}.

Despite these promising results, the open-ended nature of LLM-generated outputs has raised concerns about their robustness~\cite{huang2023}. Existing approaches for feature generation focus on either \textit{feature-feature relationships} with predefined operators~\cite{CAAFE} or \textit{feature-target relationships} with unbounded rule conditions~\cite{FeatLLM}. They rely on in-context learning by LLM with arbitrary domain knowledge and a few samples, which inherently entails risks of inconsistency and unreliability in outputs. Accordingly, evaluating the reliability of LLM-generated features remains a significant challenge.


\input{Figures/Figure_1}
\input{Figures/Figure_2}

For example, in Figure \ref{fig:motivation}, given the task of predicting diabetes for a new patient, an LLM is asked to generate a set of new features describing feature-target relationships (i.e., between patient information and diabetes). An ensemble classifier then uses these new features to make a final prediction. The LLM-generated features produced by the state-of-the-art approach~\cite{FeatLLM}, however, could be ineffective (e.g., `\texttt{Glucose} $\leq100$ and \texttt{Insulin} $\leq10$' for \texttt{Diabetes} $=$ yes) depending on the quality of input samples and the LLM's inherent knowledge. This variability can introduce noise into the resulting prediction probabilities, potentially degrading the overall classifier performance.

For more practical and reliable LLM-generated features, it is crucial to understand the consistency of their performance on feature engineering under varying contexts. Although significant strides have been made in evaluating the robustness of LLM~\cite{chang2024survey, kenthapadi2024grounding}, there remains insufficient exploration of these aspects in feature engineering, especially in the context of feature-target relationships. A recent work ELF-GYM~\cite{ELFGym} has attempted to compare LLM-generated features with human-crafted ones, but further investigation is lacking regarding the capabilities and limitations of LLMs with varying domain knowledge and examples.

To address this gap, we propose a framework that systematically diagnoses and evaluates the robustness of LLMs in feature engineering for tabular data. We focus on how consistently LLMs maintain reliability in engineering features for tabular prediction, the most prevalent task in tabular learning. Specifically, drawing inspiration from real-world practices of domain experts, we identify three core elements considered in feature engineering: golden variable, golden relation, and golden value. Our framework incorporates a novel \textit{multi-level scheme} to analyze LLM-generated features, specifically addressing the following research questions.
\vspace{-0.2cm}
 \begin{itemize}[leftmargin=10pt, noitemsep]
     \item \textbf{RQ1 (Golden Variable):} \textit{Can LLMs identify key variables highly correlated with target classes given varying domain knowledge?}
    \item \textbf{RQ2 (Golden Relation):} \textit{Can LLMs understand the causal relationship (i.e., correlation polarity) between golden variables and target classes?}
    \item \textbf{RQ3 (Golden Value):} \textit{Can LLMs set the decision boundary values of golden variables that differentiate the target classes?}
\end{itemize}

Figure \ref{fig:framework} shows the overall procedure of the proposed diagnosis and evaluation framework. Based on the multi-level scheme, we first conduct a reliability diagnosis to assess the consistency in LLM responses across varying contexts at each level. This serves as a fine-grained proxy to measure the trustworthiness of an LLM in generating features for a given dataset. The robustness of an LLM given a dataset directly influences the quality of the generated features; less robust models may produce features of varying quality, leading to prediction performance degradation. Thus, we further conduct an evaluation on the generated features to investigate how high-quality features can enhance the effectiveness of LLM-driven feature engineering and ultimately improve the prediction performance.

\paragraph{Summary} We demonstrated the efficacy of the proposed framework through comprehensive experiments on six LLMs and eight benchmark datasets
In brief, the multi-level diagnosis results show that the robustness of LLMs in feature engineering varies significantly across datasets with diverse domains, and the multi-level evaluation contributes to the improvement of few-shot prediction performance. Our key contributions and findings are summarized as follows:
 \begin{itemize}[leftmargin=10pt, noitemsep]
    \item To the best of our knowledge, this is the first work to address the robustness of LLMs in feature engineering with feature-target relationships for tabular prediction in the multi-level scheme.
    \item With reliability diagnosis, we confirm the significant variations in the robustness of LLMs in feature engineering across different datasets.
    \item Our analysis reveals that simply adding more descriptions or examples does not necessarily lead to performance gains, whereas providing high-quality examples is critical for improving the robustness of LLMs in feature engineering.
    \item We empirically demonstrate that utilizing high-quality features identified through our evaluation scheme enhances the prediction performance of the state-of-the-art method by up to $10.52\%$.
\end{itemize}

%% file: Figures/Figure_1.tex
\begin{figure}[bt!]
  \includegraphics[width=\columnwidth]{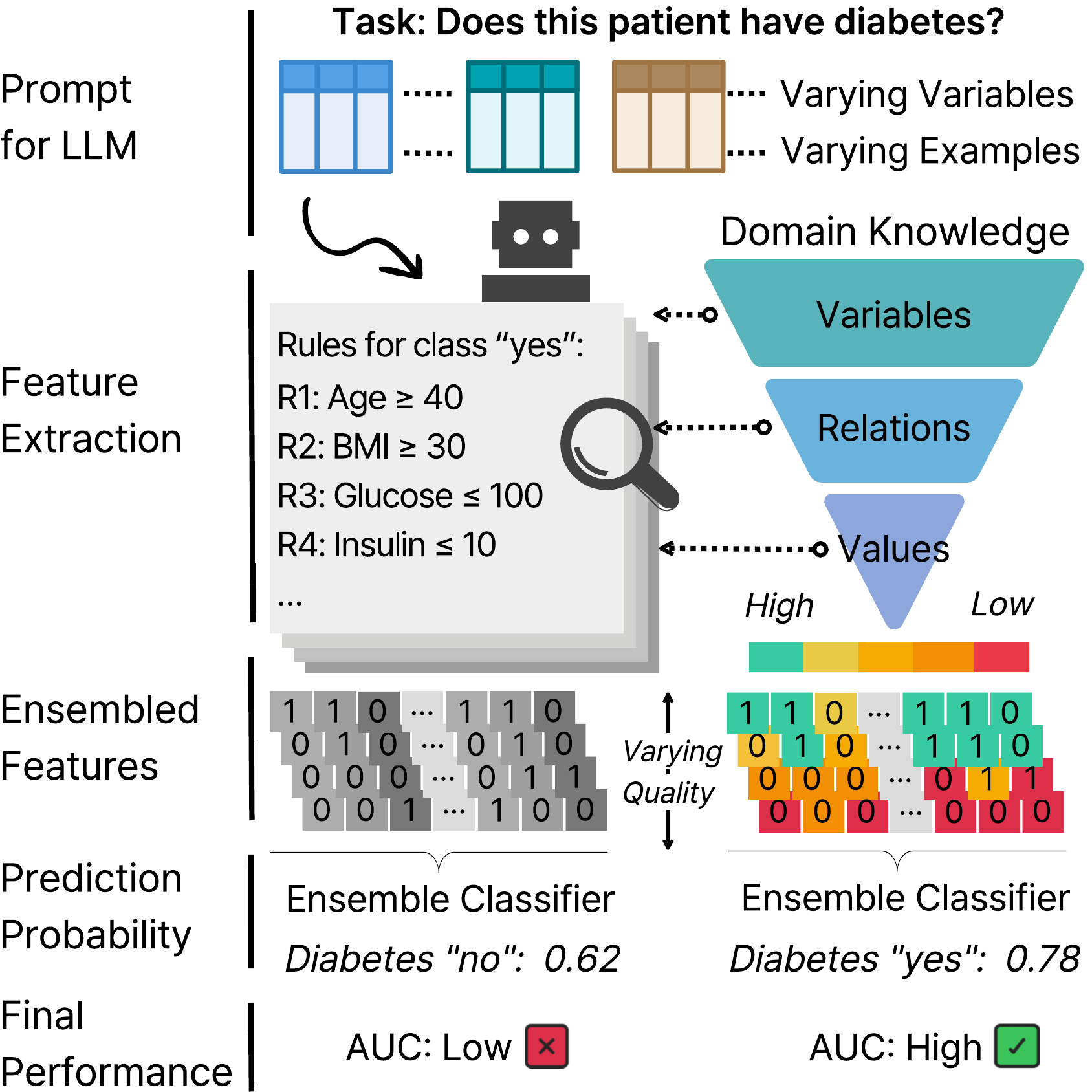}
  \caption{LLMs vulnerable to generating features of varying quality (left). Measures for high-quality features leading to performance improvement (right).}
  \vspace{-0.5cm}
  \label{fig:motivation}
\end{figure}

%% file: Figures/Figure_2.tex
\begin{figure*}[hbt!]
  \includegraphics[width=1\linewidth]{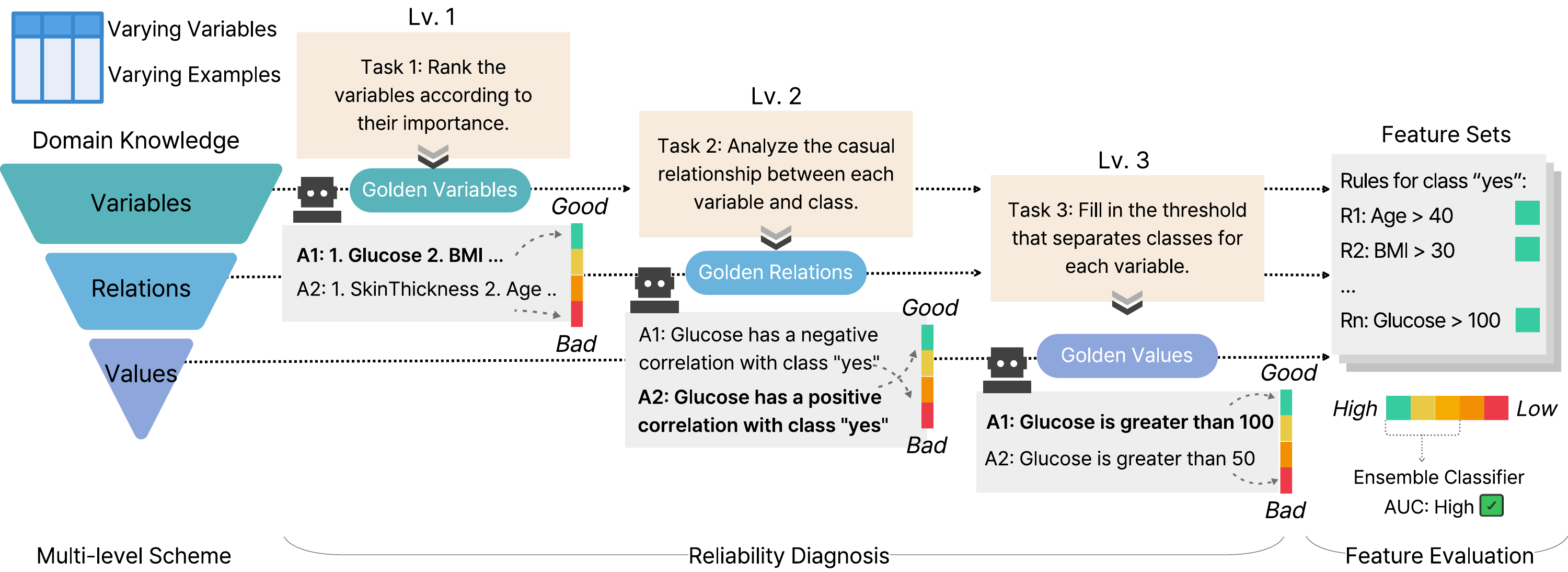} \hfill
  \vspace{-0.4cm}
  \caption {Overall procedure of our framework involves a multi-level scheme of variables, relations, and values to diagnose reliability and evaluate features generated by LLMs in feature engineering on different domains and inputs.}
  \vspace{-0.3cm}
  \label{fig:framework}
\end{figure*}

%% file: 2.__Related_Work.tex
\section{Related Work}
\label{sec:related_work}
\subsection{\chgd{{Few-shot Tabular Learning}}}
\chgd{Tabular data, consisting of distinct data instances (i.e., rows) and their variables (i.e., columns), is one of the most prevalent data types in real-world. Achieving high and robust predictive performance can provide significant benefits in data science applications~\cite{ruan2024language}. A longstanding challenge of learning on tabular data involves reasoning over structured and semantically sparse data, where each variable has a fixed type (e.g., numerical or categorical)  with potentially unbounded values and a domain-specific context within the predictive modeling task.}

\chgd{Recently, LLMs, initially trained on extensive textual corpora, have demonstrated significant capabilities in generalizing to unseen tasks~\cite{fewshotlearners}, prompting investigations into their utility for tabular data learning. Early approaches focused on converting tabular data into serialized textual prompts, enabling direct handling by LLMs~\cite{dinh2022lift, TabLLM, wang2023unipredict, wang2024meditab, zhang2023towards}. Despite its effectiveness in few-shot settings, the reliance on an expensive LLM for the entire inference process, coupled with limited interpretability, poses practical challenges. Consequently, the research focus has been shifted toward utilizing LLMs primarily as \textit{feature engineers} rather than employing them in an end-to-end, black-box prediction.}

\subsection{LLMs for Feature Engineering}
\label{subsec:feature_generation}

\paragraph{Feature Selection}
Recent studies have explored the use of LLMs for extracting domain-relevant knowledge to aid in feature selection tasks. \citealp{LMPriors2} proposed leveraging LLMs as a knowledge source to guide feature selection with the induced feature importance. Building on this idea, \citealp{LLMSelect} introduced three different prompts that directly utilize the textual outputs generated by LLMs for feature selection tasks. Additionally, \citealp{FSelect} demonstrated that this text-based approach is not only more robust than traditional data-driven approaches based on statistical inference from samples but also delivers competitive performance across diverse scenarios, including resource-limited settings.

\paragraph{Feature Generation}
\chgd{To move beyond the straightforward selection of predefined features, researchers have increasingly leveraged LLMs to generate features. A line of research focuses on \textit{feature-feature relationships}, by utilizing predefined operators (e.g., add or multiply). \citealp{CAAFE} integrated LLMs into the AutoML process to iteratively generate additional features by leveraging the dataset's semantic and contextual descriptions, enhancing model performance by embedding domain knowledge. \citealp{ELFGym} proposed a framework that assesses the quality of LLM-generated features by comparing them to human-engineered ones. They quantified the gap between the two feature sets in terms of semantic and functional similarity and identified its impact on downstream task performance.}

\chgd{On the other hand, another line of research emphasizes \textit{feature–target relationships}, aiming to generate feature-wise rules directly related to each target class. \citealp{FeatLLM} employed LLMs to create binary features through rule generation and parsing, achieving significant improvements in downstream tabular prediction tasks. However, the core challenge in this approach is that selecting and transferring meaningful rule conditions involves navigating a large combinatorial search space grounded in a given table schema and domain logic. Moreover, due to the limited input sequence lengths in LLM, tabular inference performance remains highly sensitive to subtle variations in prompts and potentially spurious correlations in samples~\cite{wen2024supervised, gardnerlarge}.}

\chgd{However, no studies to date have systematically evaluated the robustness of LLMs in feature engineering, specifically in the scope of feature-target relationships, leaving a critical gap in understanding their consistency and reliability. This gap is especially important given the complex and potentially unbounded search space for feature engineering, which often leads models to produce overly broad or unstable responses. In response, our framework focuses on analyzing the reliability of LLMs in feature–target settings, where prior work has reported strong performance but robustness has not been thoroughly studied.}

%% file: 3.__Methodology.tex
\section{Methodology}
\label{sec:methodology}
\subsection{Preliminary}
\label{subsec:preliminary}
Given a tabular dataset \( D = \{(x^i, y^i)\}_{i=1}^N \) of \( N \) labeled samples, each sample \( x^i \) includes an original feature set of \( d \) dimensional variables, \( F = \{f_j\}_{j=1}^d \). We utilize LLMs to transform the original feature set \( F \) into a new feature set \( F' \) through prompted feature engineering, $F \xrightarrow{\text{LLM}} F'$.
The transformed feature set \( F' \) is then used as input to a classifier to predict the target class \( y \).
A general pipeline of relevant works is summarized as:

\begin{enumerate}[leftmargin=12pt, noitemsep]
    \item \textbf{Prompting for LLM}: The first step involves providing a well-structured input prompt to the LLM. This input typically includes a task description, variable descriptions, and a few samples with original features and true labels.
    \item \textbf{Feature Selection/Generation via LLM}: Once prompted, LLMs either select relevant features from the dataset (\textit{feature selection}) or generate new feature representations (\textit{feature generation}). For example, an LLM-driven feature rule can transform an original feature such as \texttt{Glucose} into a new feature rule \texttt{Glucose $\geq$ 100}.
    \item \textbf{Featurization}: The next step is to transform the new feature of samples into structured input for classification modeling. For example, the new feature rules \texttt{$\{$Age $\geq$ 40, Glucose $\geq$ 100, $\ldots\}$} can form a corresponding binary feature set \textit{$[0,1,\ldots]$} of a sample. 
    \item \textbf{Model Training}: The final step involves training a machine learning model using the new feature set. This phase assesses the effectiveness of LLM-generated features by evaluating predictive performance.
\end{enumerate}

\subsection{Overview}
\label{sec:overview}
To evaluate the robustness of LLMs in feature engineering, we propose a multi-level diagnosis and evaluation framework built upon three fundamental aspects of domain expertise, which are essential for reliable feature engineering.
\begin{itemize}[leftmargin=10pt, noitemsep]
    \item \textbf{Level 1 (Identifying Key Variables):} LLMs are tested on their ability to recognize the most important variables for a given task. Domain experts can readily identify important variables that are crucial for prediction, such as \texttt{Glucose} in diabetes classification. We introduce perturbations in variable descriptions and samples to examine whether LLMs can consistently rank the correct variables.
    \item \textbf{Level 2 (Understanding Variable-Class Relationships):} This level evaluates whether LLMs can correctly determine the causal relationship between variables and target classes. While experts understand that high \texttt{Glucose} levels are positively correlated with diabetes, an LLM might generate incorrect associations depending on input variations. We test robustness by altering sample quality and variable value mixing.
    \item \textbf{Level 3 (Setting Decision Boundaries):} Domain knowledge is often reflected in the ability to determine boundary values that separate classes. For example, experts might set a \texttt{Glucose} threshold above 100 to indicate diabetes. We assess whether LLMs can provide stable decision boundaries under different input perturbations.
\end{itemize}

Based on the multi-level scheme, we first assess the reliability of LLM responses to evaluate their ability to handle variations in input conditions. This assessment helps determine the robustness of LLM-driven feature engineering across different models and datasets. Furthermore, we utilize the multi-level scheme as a framework for feature evaluation, ensuring that LLM-generated features align with domain knowledge and maintain high quality.

In this study, we demonstrate how each factor in the multi-level scheme can be derived from statistical information in datasets. In real-world scenarios, diagnosis and evaluations can be easily performed based on criteria established by domain experts.

\subsection{Multi-level Reliability Diagnosis}
At each level, we introduce variations in the input and measure how LLM-generated responses change. The variations include differences in variable descriptions, ordering, sample quality, and mixing strategies. This setup allows us to categorize LLM outputs into \textit{high-score cases}, where the predictions align with domain knowledge, and \textit{low-score cases}, where inconsistencies emerge. By analyzing the response patterns under different conditions, we gain insights into how robust LLMs are in performing feature engineering tasks.

\subsubsection{Level 1: Golden Variable}
\paragraph{Definition}
Among the variables in $F$, we define $F_{\text{golden}}$ as the subset of variables most strongly associated with the target class $y$:
\[
\small
F_{\text{golden}} = \{f_j \mid |\text{Covariance}(f_j, y)| \geq \gamma\}.
\]
Specifically, covariances between each variable and the target class are computed and ranked by their absolute values \cite{filtermethod}. The elbow method is then used to determine a threshold $\gamma$ by identifying the largest gaps. Categorical variables are represented by the one-hot encoded feature having the highest absolute covariance.

\paragraph{Prompt} An LLM is asked to rank the variables in order of importance, provided with a task description, variable descriptions, and examples:
\begin{tcolorbox}[colback=black!5,colframe=black!40,top=1mm,bottom=1mm]
\small
\textit{\ldots, rank variables according to their importance to solve the task, \ldots, [Task] [Variables] [Examples]}
\end{tcolorbox}
\noindent\chgd{The detail of information for variables and example conditions can be varied to measure reliability at level 1. See Appendix \ref{sec:apx_full_prompt} for the complete prompt.}

\paragraph{Reliability Score ($\mathcal{RS}_1$)}
 Using the rankings obtained from the LLM's responses and the identified golden variables, a rank score is computed to evaluate how well the golden variables are positioned in the higher ranks.

The rank score for each golden variable $f \in F_{\text{golden}}$ is defined as:
\[
\small
S_{Rank}(f) = 1 - \frac{\text{Rank}(f) - 1}{|F|},
\]
where $\text{Rank}(f)$ represents the rank of variable $f$ in the LLM's response, and $|F|$ is the total number of variables in the dataset.
The overall reliability score for Level 1 is calculated as the average rank score of all golden variables.

\subsubsection{Level 2: Golden Relation}
\paragraph{Definition}
\chgd{The golden relation between golden variables $F_{\text{golden}}$ and target class $y$ is defined by the direction of their correlation:}
{
\[
\small
R_{\text{golden}} = 
\begin{cases} 
\text{Positive}, & \text{if } \text{Covariance}(f, y) > 0, \\
\text{Negative}, & \text{if } \text{Covariance}(f, y) < 0.
\end{cases}
\]
}

\paragraph{Prompt} An LLM is asked to identify the relationship between key variables and target classes, provided with a task description, variable descriptions, and examples.
\begin{tcolorbox}[colback=black!5,colframe=black!40,top=1mm,bottom=1mm]
\small
\textit{\ldots, analyze the causal relationship or tendency between each variable and class, \ldots, [Task] [Variables] [Examples]}
\end{tcolorbox} 
\noindent The number of examples, sampling methods for examples, and variable corruption can be varied to measure reliability at level 2. See Appendix \ref{sec:apx_full_prompt} for the complete prompt.

\paragraph{Reliability Score ($\mathcal{RS}_2$)} 
To measure the accuracy of LLM-generated variable-class relations, we define a correctness function based on the exact match principle. Given a feature \( f \in F_{\text{golden}} \), its golden relation $R_{\text{LLM}}$ from LLM, and true golden relation \( R_{\text{golden}} \), we define a correctness score as:
\[
\small
S_{correct}(f, R_{\text{LLM}}, R_{\text{golden}}) = \mathds{1}_{(R_{\text{LLM}} = R_{\text{golden}})}.
\]

The overall reliability score for Level 2 is computed as the average correctness score across all golden variables.

\subsubsection{Level 3: Golden Value}
\paragraph{Definition}
Since domain experts typically have insights into distinguishing classes based on key variable values, we define the golden value as the specific variable value that best separates the classes. Specifically, for numerical variables $f$ with range $[f^{\min}, f^{\max}]$, the golden value is the value $v$ that maximizes the AUC score:
\[
\small
V_{\text{golden}} = \text{argmax}_{v \in [f^{\min}, f^{\max}]} \text{AUC}.
\]
For categorical variables, the golden value is the value most correlated with the target class.

\paragraph{Prompt} An LLM is asked to fill in the feature condition, provided with a task description, variable descriptions, and examples.
\begin{tcolorbox}[colback=black!5,colframe=black!40,top=1mm,bottom=1mm]
\small
\textit{\ldots, fill in the variable conditions for each class to solve the task. [Task] [Variables] [Examples]}
\end{tcolorbox} 

\noindent The number of examples, sampling methods for examples, and variable corruptions can be varied to measure reliability at level 3. See Appendix \ref{sec:apx_full_prompt} for the complete prompt.

\paragraph{Reliability Score ($\mathcal{RS}_3$)}
Given a variable \( f \in F_{\text{golden}} \), the value $V_{\text{LLM}}$ returned from LLM, and true golden value \( V_{\text{golden}} \), the correctness of predicted threshold value is evaluated using normalized error as follows:

\[
\small
\mathcal{RS}_3 = 1 - \lvert N(V_{\text{LLM}}) - N(V_{\text{golden}}) \rvert,
\]
\noindent where $N()$ is the min-max normalization. The overall reliability score is computed as the average correctness score across all golden variables.

\input{Figures/Figure_3_4}

\subsubsection{Diagnosis Result Highlights}\label{subsubsec:diagnosis_highlight}
We preview the diagnosis results before fully discussing in Section~\ref{sec:reliability_diagnosis_results}. In Figure~\ref{fig:methodology1}, the average reliability scores of models in the default setting vary across datasets. This demonstrates the uncertainty in the reliability of LLMs for feature engineering, which depends on their prior knowledge of the dataset domain. Figure~\ref{fig:methodology2} highlights how \texttt{bias} (i.e., correct responses) and \texttt{variance} (i.e., consistent responses) in an LLM's reliability fluctuate across datasets with varying inputs, emphasizing the necessity of evaluating the quality of LLM's feature engineering results.

\subsection{Multi-level Feature Evaluation}\label{subsec:multi_level_rule_evaluation}
To address the uncertainty in the robustness of LLMs on different datasets, we introduce a simple yet effective method for verifying the quality of transformed feature set $F'$ through the multi-level evaluation scheme. The corresponding results and analyses, examined using the state-of-the-art feature engineering method \cite{FeatLLM}, are presented in Section~\ref{sec:feature_evaluation}.

\subsubsection{Level 1: Golden Variable}
\paragraph{Feature Score ($\mathcal{FS}_1$)}
To evaluate the correctness of feature selection, we measure the F1-score of the variables in the transformed feature set $F_{\text{LLM}}$ against $F_{\text{golden}}$:
\[
\small
\mathcal{FS}_1 = \frac{2 \times P \times R}{P + R} \text{, where}
\]
\[
\small
P = \frac{\sum_{f \in F_{\text{LLM}}} S_{correct}(f, F_{\text{LLM}}, F_{\text{golden}})}{|F_{\text{LLM}}|}
\]
\[
\small
R = \frac{\sum_{f \in F_{\text{golden}}} S_{correct}(f, F_{\text{LLM}}, F_{\text{golden}})}{|F_{\text{golden}}|}
\]

\subsubsection{Level 2: Golden Relation}
\paragraph{Feature Score ($\mathcal{FS}_2$)}

Transformed feature sets are evaluated based on their alignment with the class-specific variable relations. Given a variable \( f \), LLM-generated relation \( R_{\text{LLM}} \), and ground-truth relation \( R_{\text{golden}} \), the overall score for golden relation evaluation is defined as:
\[
\small
\mathcal{FS}_2 = \frac{1}{|F_{\text{golden}}|} \sum_{f \in F_{\text{golden}}} S_{correct}(f, R_{\text{LLM}}, R_{\text{golden}}).
\]

\subsubsection {Level 3: Golden Value}
\paragraph{Feature Score ($\mathcal{FS}_3$)}

The correctness of predicted threshold value is evaluated using normalized error as follows:
{
\[
\small
\mathcal{FS}_3 = 1 - \lvert N(V_{\text{LLM}}) - N(V_{\text{golden}}) \rvert,
\]
}
\noindent where $N()$ is the min-max normalization.
For categorical variables, $\mathcal{FS}_3 = 1$ if value $\in V_{\text{golden}}$ or 0.5 otherwise.

%% file: Figures/Figure_3_4.tex
\begin{figure*}[t!]
  \centering
  \begin{minipage}{0.49\textwidth}
  \centering
\includegraphics[width=0.72\columnwidth]{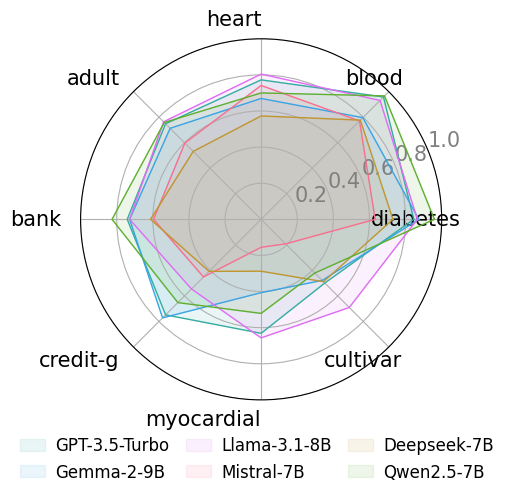}  
  \caption{Variation of reliability scores (averaged over three levels) for different LLMs and datasets.}
  \label{fig:methodology1}
  \end{minipage}
    \hfill
    \begin{minipage}{0.49\textwidth}
\includegraphics[width=\columnwidth]{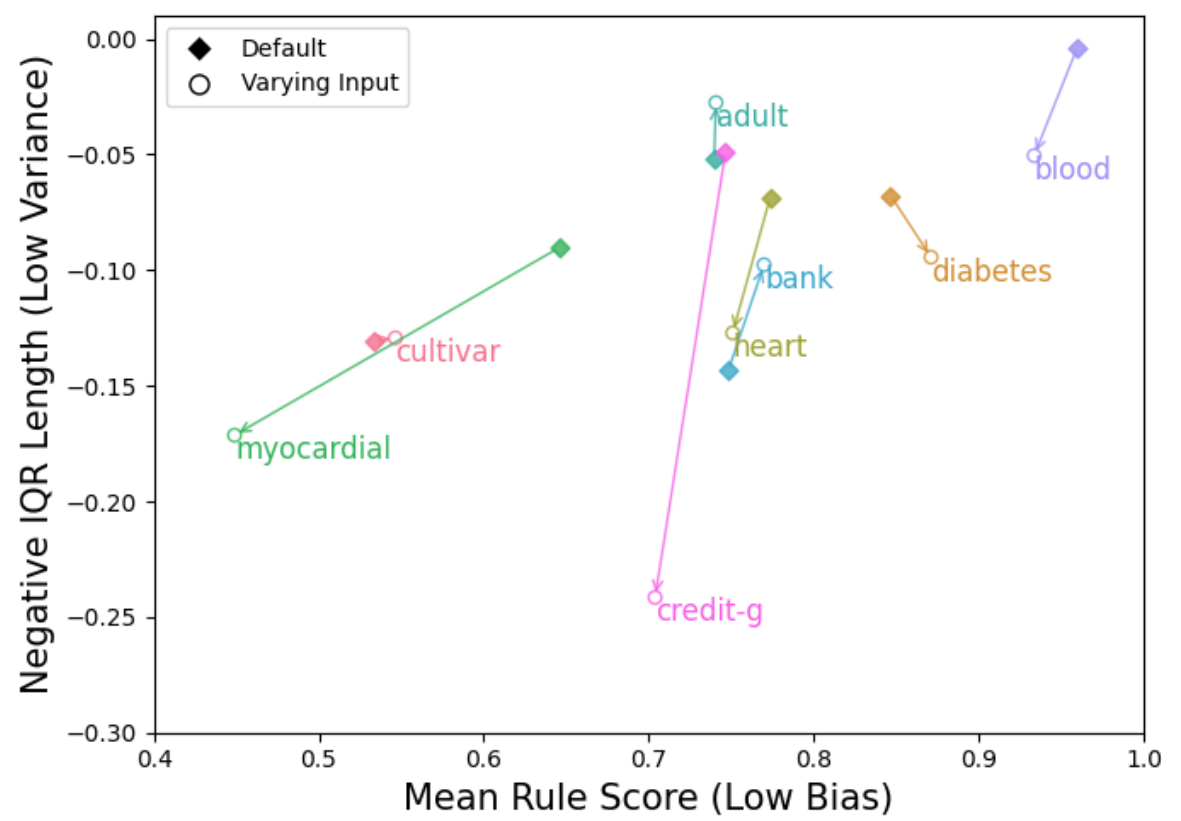}
  \caption{\chgd{The change of variance and bias of reliability score with varying input for GPT-3.5-Turbo.}}
  \label{fig:methodology2}
  \end{minipage}
  \vspace{-0.5cm}
\end{figure*}

%% file: 4.__Experiments.tex
\section{Experiments}\label{sec:experiments}
\subsection{Reliability Diagnosis}
\label{sec:reliability_diagnosis_results}
\subsubsection{Reliability Diagnosis Setting}
\label{subsec:reliability_diagnosis_setting}
\paragraph{LLMs}
We employ \texttt{GPT-3.5-Turbo} as the base model, following the state-of-the-art feature engineering method \cite{FeatLLM}. Considering the versatility and usability in various scenarios, we also employ lightweight models, which are \texttt{Gemma-2-9B, Llama-3.1-8B, Mistral-7B, Qwen2.5-7B, Deepseek-7B} to compare their reliability scores with those of the base model.

\input{Figures/Figure_5}

\paragraph {Datasets}
We utilized eight binary classification datasets commonly adopted in recent studies on tabular feature engineering and prediction, ensuring consistency with prior research. These datasets were selected based on several criteria, including diverse application domains—such as healthcare (e.g., Blood~\cite{blood}, Diabetes~\cite{Diabetes}, Heart~\cite{Heart}), finance (e.g., Adult~\cite{adult}, Bank~\cite{bank}, Credit-g~\cite{credit-g}), and agriculture (e.g., Cultivar~\cite{Cultivar})—as well as varying dataset sizes (e.g., Myocardial~\cite{myocardial}, comprising 111 variables). Additionally, we incorporated an out-of-distribution scenario using Cultivar, representing a domain that LLMs are unlikely to have encountered during pre-training. 

\subsubsection{Reliability Diagnosis Results}\label{sec:diagnosis}
Figure \ref{fig:reliaility_score_distribution} shows the reliability diagnosis results of each level across models and datasets. To further understand the factors that affect reliability scores, we conduct a series of analyses at each level. (Figures \ref{fig:desc_vs_example}--\ref{fig:shot_vs_sampling} show representative results; see Appendix \ref{apx:reliability_detail} for the full results).

\paragraph{Which Model Performs Best in the Zero-shot Setting?}
We compare LLMs by providing detailed variable descriptions without examples across levels. Even within the same dataset, models exhibit distinct strengths at each level. GPT-3.5-Turbo outperforms in identifying golden variables at Level 1, particularly excelling in Credit-g. However, its performance drops significantly in Level 2, where it must infer relationships between variables and the target class. Conversely, Gemma-2-9B shows strong performance in Credit-g at Levels 2 and 3. Llama-3.1-8B excels in Level 2, especially in Cultivar, but performs worse in Bank compared to other levels. Qwen-2.5-7B consistently performs well across all three tasks in Blood and Diabetes, though it exhibits large performance gaps across levels in Bank and Adult. Deepseek-7B performs the worst, often responding with “neutral” when asked about correlations in Level 2. Mistral-7B also struggles in Level 2 and 3, producing hallucinated responses when faced with datasets containing many variables (e.g., Myocardial) or out-of-distribution scenarios (e.g., Cultivar).

To analyze how different input conditions affect LLM responses, we conduct experiments using GPT-3.5-Turbo under controlled variations in the following robustness analyses. Specifically, we further adjusted the number of shots, sampling methods, and variable corruption strategies.

\input{Figures/Figure_6}
\input{Figures/Figure_7}
\paragraph{Does Adding More Descriptions and Examples Improve Robustness?}
Figure \ref{fig:desc_vs_example} compares reliability scores in terms of \texttt{bias} when either descriptions or examples are added under the simplest description setting (i.e., variable name only) in zero-shot. Across most datasets (e.g., Adult, Bank, Blood, Cultivar, Heart), adding examples improves the reliability score more effectively than descriptions. In some cases (e.g., Blood and Cultivar), additional descriptions even degrade performance. For datasets with a large number of variables, such as Myocardial, both descriptions and examples negatively impact the reliability score.
These findings indicate that additional information does not always lead to better outcomes.
Figure \ref{fig:reliability_score_change} shows the different types of robustness change patterns by dataset, when few-shot samples are given with detailed variable descriptions.
\vspace{0.2cm}
\begin{itemize}[leftmargin=10pt, noitemsep]
    \item Negative Impact of Few-Shot: In some cases, the few-shot approach produced lower scores than zero-shot. For example, in the credit-g dataset, few-shot resulted in lower scores and greater variability (Level 1). Similar patterns were observed in the myocardial, credit-g, and blood datasets (Level 2). In myocardial, heart, and blood, adding sample data did not help and sometimes led to decreased scores (Level 3). This finding implies that the provided samples sometimes functioned as noise.
    \item No Effect: Several datasets showed little to no difference between zero-shot and few-shot performance. For instance, adult, diabetes, and myocardial showed no significant change (Level 1). Adult and cultivar maintained stable performance across both methods (Level 2). Bank did not benefit from the sampling approach, showing no notable score changes (Level 3). This indicates that methods other than few-shot prompting might be required to help the model learn meaningful variable–target relations.
    \item Dependence on Sampling: Some datasets benefited from few-shot prompting, but the effectiveness depended heavily on the sample quality. For example, bank, blood, cultivar, and heart showed performance improvements (Level 1). For bank, diabetes, and heart, the presence of samples improved scores, albeit inconsistently depending on the sampling method (Level 2). Datasets such as diabetes, adult, credit-g, and cultivar exhibited improved scores when samples were provided, although credit-g was particularly sensitive to the quality of those samples (Level 3). This underscores the importance of sample selection that aids the model’s reasoning.
    \end{itemize}
    
\input{Figures/Figure_8}

\paragraph{Does Sampling Matter More than the Number of Shots?}
Figure \ref{fig:shot_vs_sampling} shows the impact of the number of shots and the sampling method at Levels 2 and 3. Across most datasets, sampling quality exerts a stronger influence on robustness than the number of shots, with larger effects observed in Heart and Cultivar at Level 2 and in Credit-g and Bank at Level 3. Besides, in-depth analysis between worst/best sampling and random sampling reveals additional dataset-specific patterns. At Level 1, low-quality examples in Bank substantially degraded performance when identifying key variables. At Level 2, high-quality samples in Credit-g led to significant gains when inferring variable–class relations. At Level 3 the gap between high- and low-quality examples is substantial when identifying golden values. In Bank and Blood, however, this gap decreases as the number of shots increases, indicating that more examples can mitigate the negative effect of poor-quality examples.

\input{Tables/Table_1}
\subsection{Feature Evaluation}
\label{sec:feature_evaluation}
\subsubsection{Feature Evaluation Setting}
\label{subsec:feature_evaluation_setting}
We evaluate the performance of binary classification in the eight datasets. We also used Communities~\cite{redmond2009communities} to show extendability of our framework in a multi-class classification task (see Appendix \ref{sec:apx_communities}). We compare three conventional classifiers, (1) Logistic regression (LogReg), (2) XGBoost~\cite{xgboost}, and (3) RandomForest~\cite{randomforest}, and the state-of-the-art feature engineering method FeatLLM~\cite{FeatLLM} ensembling with ten feature sets. The primary aim of our study is to evaluate the robustness and reliability of LLM-generated features, particularly in the SOTA setting proposed by FeatLLM. Therefore, we employed our framework built upon FeatLLM, which has demonstrated superior performance over other baselines (e.g., TabLLM~\cite{TabLLM}).
In our framework, we simply averaged the Levels 1--3 evaluation scores of each feature set and selected the
top 3 or excluded the bottom 3 feature sets out of the ten feature sets. As in the prior work, these feature sets were ensembled with equal weights for the final prediction.

\subsubsection{Feature Evaluation Result}

Table \ref{tab:performance} summarizes the average AUC scores and standard deviations obtained from three runs—each with a different random seed for the training set selection—across all datasets. We observe that our framework ranks as the top performer on non-robust datasets when compared with FeatLLM. These results highlight the efficacy of our approach in scenarios where the model is particularly sensitive to input variations.
In Appendix \ref{sec:apx_feature_eval_results}, we further discuss the relationship between reliability diagnosis results and feature evaluation results.

%% file: Figures/Figure_5.tex
\begin {figure*}[hbt!]
  \begin{center}\text{\hspace{0.5cm} \small Level 1 \hspace{4.2cm} Level 2 \hspace{4.3cm} Level 3} \end{center}
  \includegraphics [width=\textwidth]{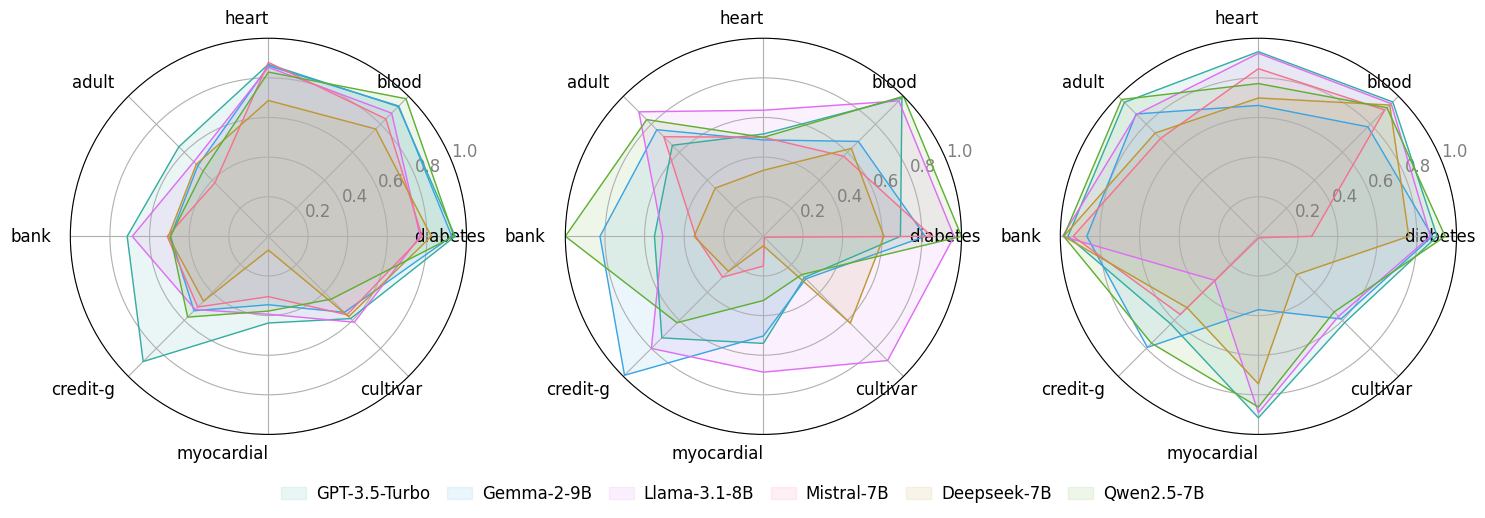}
  \caption{Variation of reliability scores of each level for different models and datasets.}
  \label {fig:reliaility_score_distribution}
  \vspace{-0.5cm}
\end {figure*}

%% file: Figures/Figure_6.tex
\begin {figure}[t!]
  \includegraphics [width=\columnwidth]{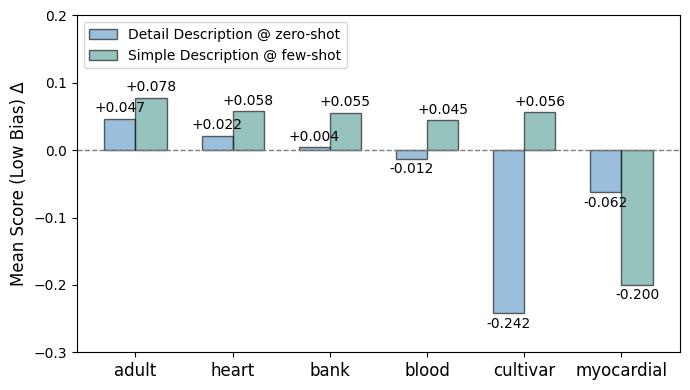}
  \caption{Impact of additional descriptions and examples on reliability scores at Level 1.}
  \label {fig:desc_vs_example}
  \vspace{-0.5cm}
\end {figure}

%% file: Figures/Figure_7.tex
\begin {figure*}[hbt!]
  \includegraphics [width=\textwidth]{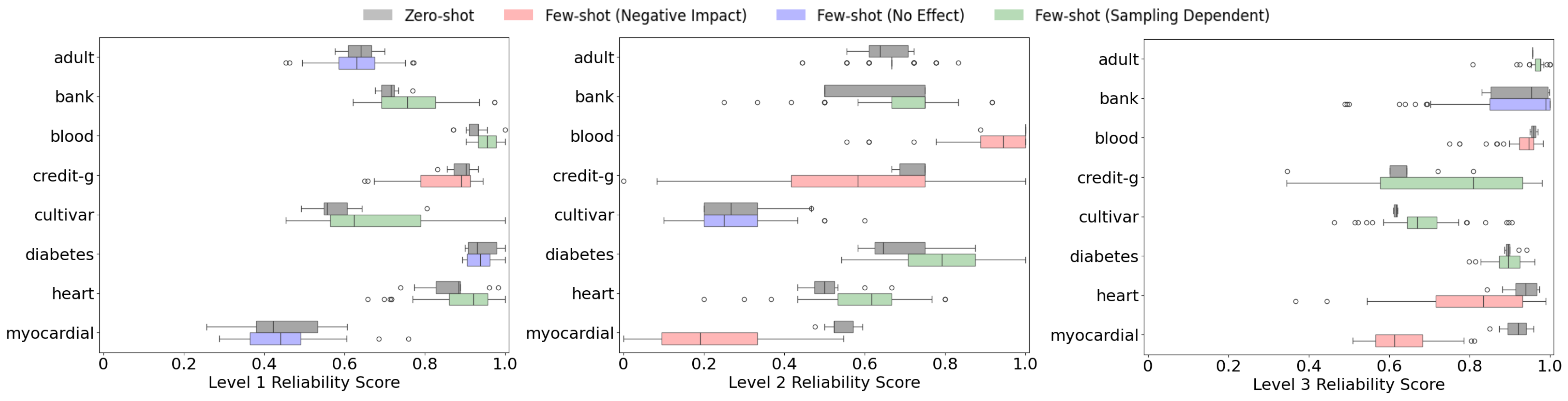}
  \caption {\chgd{Effects of varying the number of examples on reliability scores at each level for GPT-3.5-Turbo.}}
  \label {fig:reliability_score_change}
  \vspace{-0.4cm}
\end {figure*}

%% file: Figures/Figure_8.tex
\begin {figure}[t!]
  \includegraphics [width=\columnwidth]{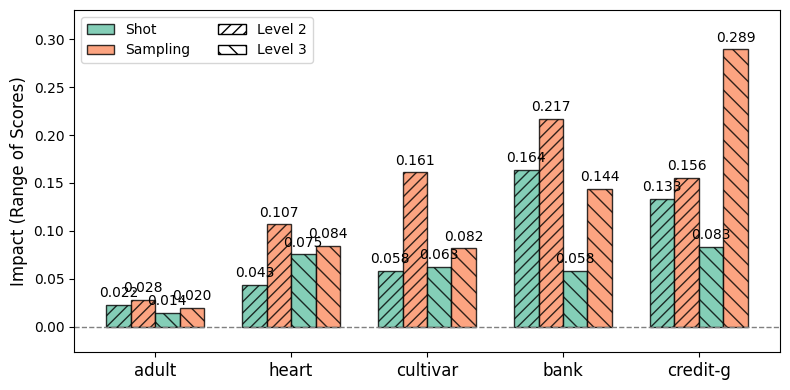}
  \caption{Impact of the number of shots and example quality on reliability scores at Levels 2 and 3.}
  \label {fig:shot_vs_sampling}
  \vspace{-0.6cm}
\end {figure}

%% file: Tables/Table_1.tex
\begin{table*}
\centering
\caption{Few-shot classification performance evaluation results. We used only the top 3 feature sets or excluded the bottom 3 feature sets based on the average evaluation scores of three levels.}
\label{tab:performance}
\resizebox{\textwidth}{!}{
\begin {tabular}{c|c|ccc|cccc}
\toprule
\ {Data}           & Shot                   & LogReg           & RandomForest          & XGBoost     & FeatLLM          & Ours (Top 3)       & Ours (w/o Bottom 3) & Improvement (\%) \\
\midrule
\multirow{3}{*}{ Credit-g}      
                            & 4                      & $\underline{56.77\pm11.93}$ & $51.35\pm8.5$ & $50.0\pm0.0$ & $52.27\pm8.38$ & $\bold{57.77\pm5.37}$ & $55.65\pm7.08$ & $\textcolor{red}{\blacktriangle} 10.52$\\ 
                            & 8                      & $49.7\pm12.84$ & $57.06\pm8.59$ & $49.84\pm6.37$ & $58.87\pm4.69$ & $\bold{62.89\pm5.72}$ & $\underline {61.49\pm7.65}$ & $\textcolor{red}{\blacktriangle} 6.83$\\ 
                            & 16                      & $\bold{64.48\pm9.71}$ & $\underline{64.27\pm11.32}$ & $59.49\pm10.36$ & $56.47\pm4.51$ & $57.51\pm1.87$ & $58.74\pm8.04$ & $\textcolor{red}{\blacktriangle} 4.02$\\ 
\midrule
\multirow{3}{*}{Myocardial}      
                            & 4                      & $54.28\pm5.09$ & $\bold{57.93\pm2.64}$ & $50.0\pm0.0$ & $54.08\pm3.28$ & $\underline{56.35\pm12.34}$ & $55.46\pm4.77$ & $\textcolor{red}{\blacktriangle} 4.20$\\ 
                            & 8                      & $5\underline{4.25\pm8.33}$ & $52.78\pm2.67$ & $\bold{55.44\pm5.34}$ & $51.6\pm7.06$ & $54.04\pm6.16$ & $52.26\pm7.69$ & $\textcolor{red}{\blacktriangle} 4.73$\\ 
                            & 16                      & $56.39\pm5.57$ & $50.96\pm5.98$ & $55.21\pm5.96$ & $58.54\pm1.84$ & $\bold{61.96\pm3.64}$ & $\underline{60.92\pm2.09}$ & $\textcolor{red}{\blacktriangle} 5.84$\\
\midrule
\multirow{3}{*}{Cultivar}      
                            & 4                      & $41.93\pm9.19$ & $44.14\pm4.23$ & $50.0\pm0.0$ & $\bold{55.84\pm4.99}$ & $55.14\pm6.45$ & $\underline{55.63\pm8.79}$ & $\textcolor{blue}{\blacktriangledown} 0.38$\\ 
                            & 8                      & $48.67\pm7.27$ & $49.2\pm4.68$ & $48.44\pm1.56$ & $56.95\pm3.52$ & $\bold{60.43\pm6.79}$ & $\underline{57.45\pm5.37}$ & $\textcolor{red}{\blacktriangle} 6.11$\\ 
                            & 16                      & $53.86\pm8.89$ & $50.28\pm5.77$ & $57.08\pm5.59$ & $\underline{57.57\pm2.67}$ & $57.49\pm3.22$ & $\bold{58.3\pm2.46}$ & $\textcolor{red}{\blacktriangle} 1.27$\\
\midrule
\multirow{3}{*}{Bank}      
                            & 4                      & $67.65\pm16.53$ & $64.28\pm5.0$ & $50.0\pm0.0$ & $74.34\pm1.71$ & $\underline{75.17\pm1.6}$ & $\bold{76.07\pm2.87}$ & $\textcolor{red}{\blacktriangle} 2.33$\\ 
                            & 8                      & $75.05\pm1.57$ & $63.36\pm7.13$ & $58.52\pm10.73$ & $76.09\pm2.57$ & $\underline{77.87\pm0.38}$ & $\bold{78.03\pm1.66}$ & $\textcolor{red}{\blacktriangle} 2.55$\\ 
                            & 16                      & $77.6\pm2.18$ & $77.69\pm2.51$ & $68.75\pm10.87$ & $\underline{79.57\pm1.01}$ & $\bold{79.59\pm2.72}$ & $79.5\pm2.96$ & $\textcolor{red}{\blacktriangle} 0.03$\\                     
\midrule
\multirow{3}{*}{Heart}      
                            & 4                      & $52.19\pm1.59$ & $\bold{79.92\pm7.71}$ & $50.0\pm0.0$ & $73.82\pm6.06$ & $\underline{77.69\pm2.7}$ & $77.18\pm3.53$ & $\textcolor{red}{\blacktriangle} 5.24$\\ 
                            & 8                      & $60.86\pm8.74$ & $\bold{81.84\pm2.88}$ & $53.76\pm11.81$ & $70.88\pm13.15$ & $\underline{76.9\pm7.8}$ & $70.99\pm10.31$ & $\textcolor{red}{\blacktriangle} 8.49$\\ 
                            & 16                      & $65.45\pm13.36$ & $\bold{85.5\pm2.39}$ & $82.99\pm1.69$ & $80.31\pm7.69$ & $\underline{83.57\pm9.29}$ & $81.08\pm5.33$ & $\textcolor{red}{\blacktriangle} 4.06$\\

\midrule
\multirow{3}{*}{Diabetes}      
                            & 4                      & $47.04\pm12.37$ & $56.67\pm11.65$ & $50.0\pm0.0$ & $79.55\pm0.35$ & $\underline{79.65\pm0.97}$ & $\bold{79.74\pm0.5}$ & $\textcolor{red}{\blacktriangle} 0.24$\\ 
                            & 8                      & $52.73\pm5.8$ & $64.19\pm6.21$ & $39.2\pm14.42$ & $\bold{80.48\pm0.21}$ & $79.71\pm0.24$ & $\underline{80.41\pm0.76}$ & $\textcolor{blue}{\blacktriangledown} 0.09$\\ 
                            & 16                      & $64.78\pm14.34$ & $67.3\pm6.02$ & $72.69\pm2.33$ & $79.85\pm0.83$ & $\bold{80.94\pm2.11}$ & $\underline{80.25\pm1.52}$ & $\textcolor{red}{\blacktriangle} 1.37$\\   
\midrule
\multirow{3}{*}{Blood}      
                            & 4                      & $42.75\pm16.56$ & $48.66\pm12.56$ & $50.0\pm0.0$ & $\bold{56.34\pm6.66}$ & $54.57\pm10.59$ & $\underline{55.89\pm6.51}$ & $\textcolor{blue}{\blacktriangledown} 0.80$\\ 
                            & 8                      & $60.27\pm8.9$ & $57.67\pm8.98$ & $55.87\pm5.1$ & $\underline{66.63\pm0.69}$ & $62.28\pm7.24$ & $\bold{66.71\pm0.84}$ & $\textcolor{red}{\blacktriangle} 0.12$\\ 
                            & 16                      & $\bold{68.59\pm3.81}$ & $51.9\pm8.84$ & $63.43\pm8.09$ & $67.61\pm1.9$ & $\underline{67.98\pm0.31}$ & $67.08\pm1.71$ & $\textcolor{red}{\blacktriangle} 0.55$\\ 
\midrule
\multirow{3}{*}{Adult}      
                            & 4                      & $58.3\pm7.89$ & $70.28\pm5.32$ & $50.0\pm0.0$ & $\bold{87.58\pm0.29}$ & $86.48\pm1.21$ & $\underline{87.55\pm0.83}$ & $\textcolor{blue}{\blacktriangledown} 0.03$\\ 
                            & 8                      & $58.97\pm8.93$ & $57.27\pm21.03$ & $59.19\pm7.96$ & $\bold{87.29\pm0.31}$ & $86.35\pm0.3$ & $\underline{86.95\pm0.15}$  & $\textcolor{blue}{\blacktriangledown} 0.39$\\ 
                            & 16                      & $67.61\pm10.76$ & $77.93\pm2.79$ & $68.17\pm9.31$ & $\underline{87.59\pm0.9}$ & $85.53\pm1.74$ & $\bold{87.61\pm0.97}$ & $\textcolor{red}{\blacktriangle} 0.02$\\ 
\bottomrule
\end{tabular}
}
\end{table*}

%% file: 5.__Conclusion.tex
\vspace{0.44cm}
\section{Conclusion}
We present a multi-level framework for evaluating the robustness of LLMs in tabular feature engineering. Our analysis reveals that the few-shot prediction performance of LLMs varies significantly across different datasets, highlighting the need for consistent and reliable methods in real-world applications. By focusing on golden variables, relations, and values, we demonstrate that high-quality features generated by LLMs can lead to substantial performance improvements. Our findings emphasize the importance of robustness in LLM-driven feature engineering and provide valuable insights for enhancing its reliability and effectiveness.

%% file: 6.__Appendix.tex
\section {Implementation Details}
\subsection {Datasets}
\label {sec:apx_dataset}
Table \ref{tab:dataset_stats} shows the basic information of each dataset used in our experiments. The numbers in parentheses under \# of features represent the number of categorical and numerical features, respectively. Similarly, the numbers in parentheses under \# of golden features represent the number of categorical and numerical golden features.
\input {Tables/Table_2}

\subsection{LLMs and Baselines}
For various LLM backbones, the temperature for LLM inference is set to nonzero (i.e., 0.5). For experiments
involving open-source models, we use vLLM 0.9.2 \cite{vllm} with two A6000 GPUs. We vary the data availability to conduct evaluations with 4-shot, 8-shot, and 16-shot configurations. The test performance is measured using a logistic/linear regression model, selected via grid search with 5-fold cross-validation. To evaluate classification tasks, we use the area under the ROC curve (AUROC) as the primary metric. 

The number of conditions included in the feature rule is determined as:
\[
\small
\max(\text{golden variables}, \text{variables} \times 0.5).
\]
This ensures a balance between model interpretability and robustness.

\subsection {Sampling}
Examples with varying levels of quality are used to evaluate the robustness of LLM responses. 
The examples provided to the LLM can act as either informative signals or noise. To evaluate how robust the LLM's prior knowledge is, we modify the quality of the provided examples and conduct experiments. Sampling is divided into best-case and worst-case scenarios based on the distance between each sample's feature and the golden value $V_{\text{golden}}$. 

\[
\small
N(V_{f}) = \frac{V_{f} - f^{\min}}{f^{\max} - f^{\min}}.
\]

When $R_{\text{golden}}$ is positive and $N(V_{f}) > N(V_{\text{golden}})$, the distance is defined as:
{\small
\[
\text{distance} = \lvert N(V_{f}) - N(V_{\text{golden}}) \rvert.
\]}

When $N(V_{f}) \leq N(V_{\text{golden}})$, the distance incorporates a penalty:
{\small
\[
\text{distance} = \lvert N(V_{f}) - N(V_{\text{golden}}) \rvert + \text{penalty}_{\text{pos}}.
\]}

The penalty is defined as:
{\small
\[
\text{penalty}_{\text{pos}} = \lvert N(f^{\max}) - N(V_{\text{golden}}) \rvert.
\]}

\section {Additional Results}
\label {sec:apx_full_result}
\subsection{Feature Evaluation Results}
\label{sec:apx_feature_eval_results}
\input{Figures/Figure_9_10}

Figures \ref{fig:robustness_scatter} and \ref{fig:auc_difference} indicate the correlation between robustness and overall performance. For datasets that exhibit large performance fluctuations depending on the input, adopting well-designed ensemble rules can lead to notable improvements. This underscores the importance of diagnosing which inputs serve as genuine ``information” as opposed to mere ``data.” Our findings demonstrate that domain knowledge serves as a valuable guide for pinpointing critical variables and mitigating irrelevant complexity. In practice, diagnosing before fully trusting a model—by uncovering its weaknesses and evaluating its robustness—offers a principled approach to optimizing performance. By strategically combining diagnostic insights with expert knowledge, practitioners can effectively enhance LLM reliability and achieve more consistent results across a range of datasets.
In Figure \ref{fig:feature_score_auc_relationship}, we further demonstrate the relationships between the multi-level evaluation scores and AUC scores across datasets.

\input {Figures/Figure_A2}
\input {Figures/Figure_A6}
\input {Tables/Table_4.tex}

\subsection{Results on the Communities Dataset}
\label{sec:apx_communities}
\chgd{For a multi-class classification setting, we report additional results of GPT-3.5-Turbo on the Communities dataset (Table \ref{tab:performance_communities}), reliability diagnosis results (Figures \ref{fig:experiments_communities1} and \ref{fig:experiments_communities2}), and performance improvement results with varying shots (Figure \ref{fig:experiments_communities3}).}
\input {Figures/Figure_A7}
\subsection {Box Plots of Reliability Diagnosis Score}\label{apx:reliability_detail}
We include additional reliability diagnosis results of GPT-3.5-Turbo (Figures \ref{fig:adult}--\ref{fig:myocardial}) across datasets and levels.
\input {Figures/Figure_A3}

\section {Prompt Examples}
\label{sec:apx_full_prompt}
In Figures \ref{fig:prompt1}--\ref{fig:prompt3}, we include example prompts designed for our multi-level reliability diagnosis in Heart dataset. In Figure \ref{fig:prompt4}, we also include an example prompt for rule evaluation proposed by FeatLLM \cite{FeatLLM}.
\input {Figures/Prompt_A1}
\input {Figures/Prompt_A2}
\input {Figures/Prompt_A3}
\input {Figures/Prompt_A4}

%% file: Tables/Table_2.tex
\begin{table}[h]
\centering
\caption{\chgd{Dataset statistics.}}
\begin{adjustbox}{max width=\columnwidth}
\begin{tabular}{l|cccc}
\toprule
Data & \# of samples & \# of features & Label ratio (\%) & \# of golden features \\
\midrule
Adult & 48842 & 14 (7/7) & 76:24 & 3 (2/1) \\
Bank & 45211 & 16 (8/8) & 88:12 & 2 (1/1) \\
Blood & 748 & 4 (0/4) & 76:24 & 3 (0/3) \\
Communities & 1994 & 103 (1/102) & 34:33:33 & 19 (19/0) \\
Credit-g & 1000 & 20 (12/8) & 70:30 & 2 (1/1) \\
Cultivar & 320 & 10 (3/7) & 50:50 & 2 (1/1) \\
Diabetes & 768 & 8 (0/8) & 65:35 & 4 (0/4) \\
Heart & 918 & 11 (4/7) & 45:55 & 5 (3/2) \\
Myocardial & 1700 & 111 (94/17) & 22:78 & 7 (1/6) \\
\bottomrule
\end{tabular}
\end{adjustbox}
\label{tab:dataset_stats}
\end{table}

%% file: Figures/Figure_9_10.tex
\begin {figure}[t!]
  \includegraphics [width=\columnwidth]{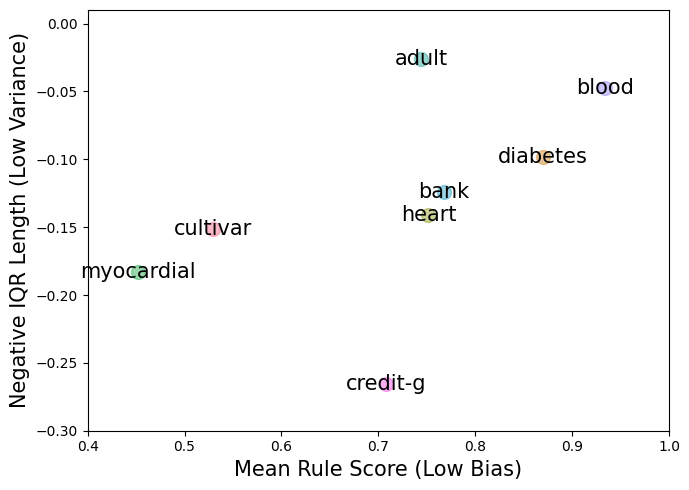}
  \caption {Variance and bias of average reliability score of GPT-3.5-Turbo.}
  \label {fig:robustness_scatter}
  \includegraphics [width=\columnwidth]{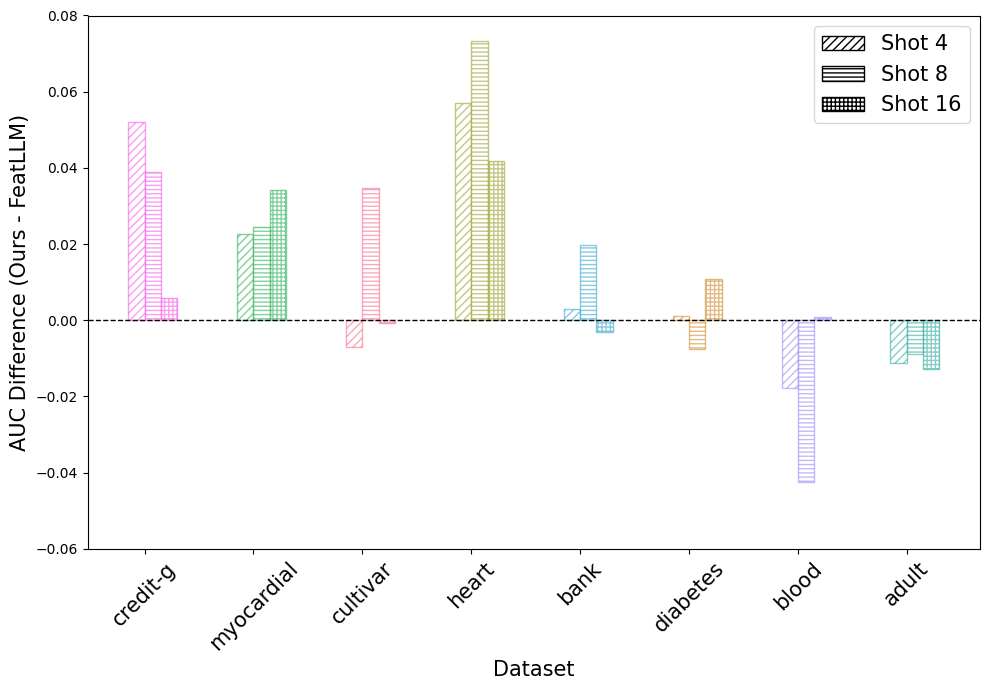}
  \caption {Performance improvement of ours over FeatLLM with different numbers of examples.}
  \label {fig:auc_difference}
\end {figure}

%% file: Figures/Figure_A2.tex
\begin {figure*}[t!]
  \vspace{-0.5cm}
  \includegraphics [width=\textwidth]{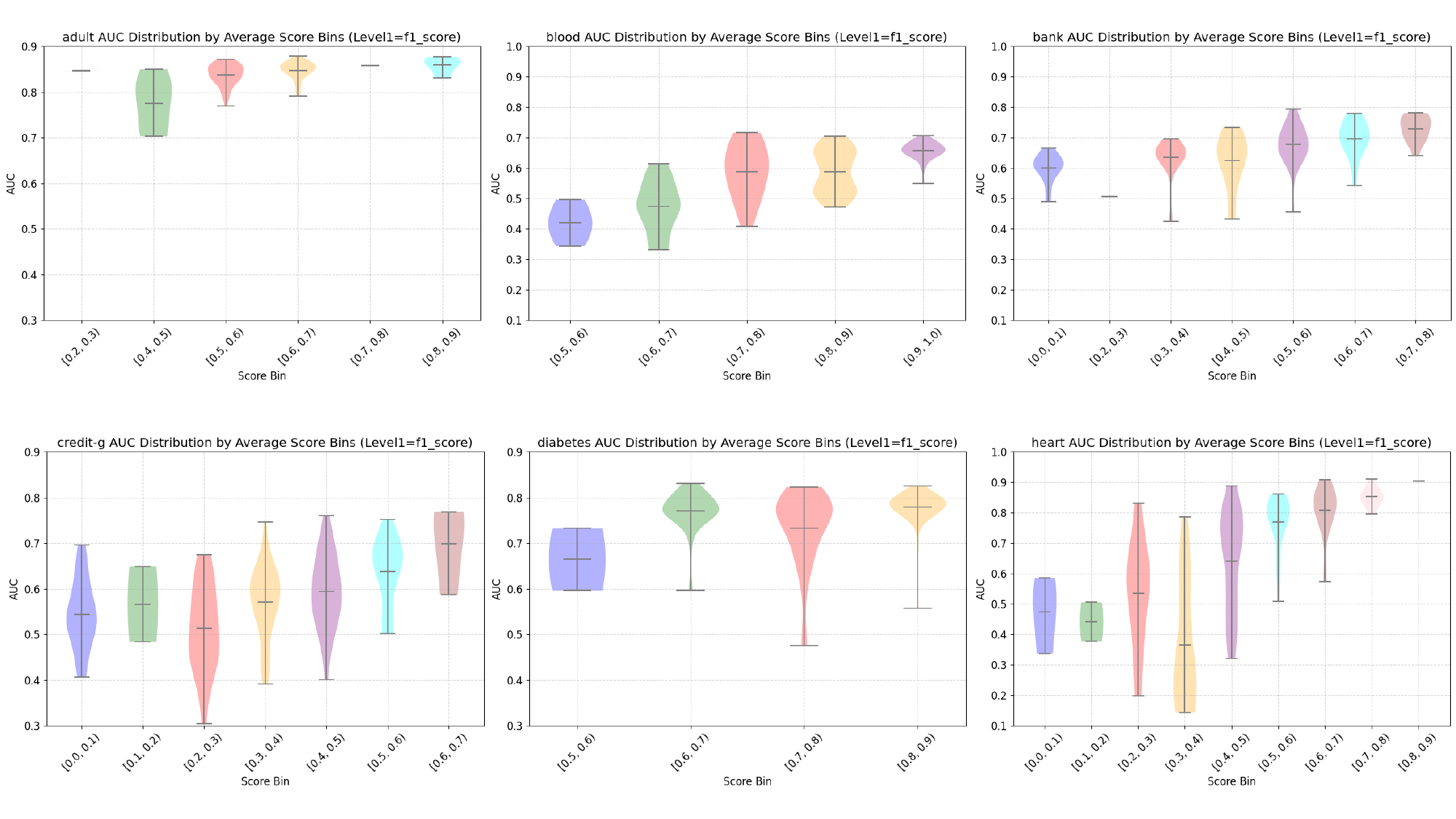}
  \caption {Relationship between feature evaluation score and AUC score.}
  \label {fig:feature_score_auc_relationship}
    \vspace{-0.5cm}

\end {figure*}

%% file: Figures/Figure_A6.tex
\begin {figure*}[t!]
  \includegraphics [width=\textwidth]{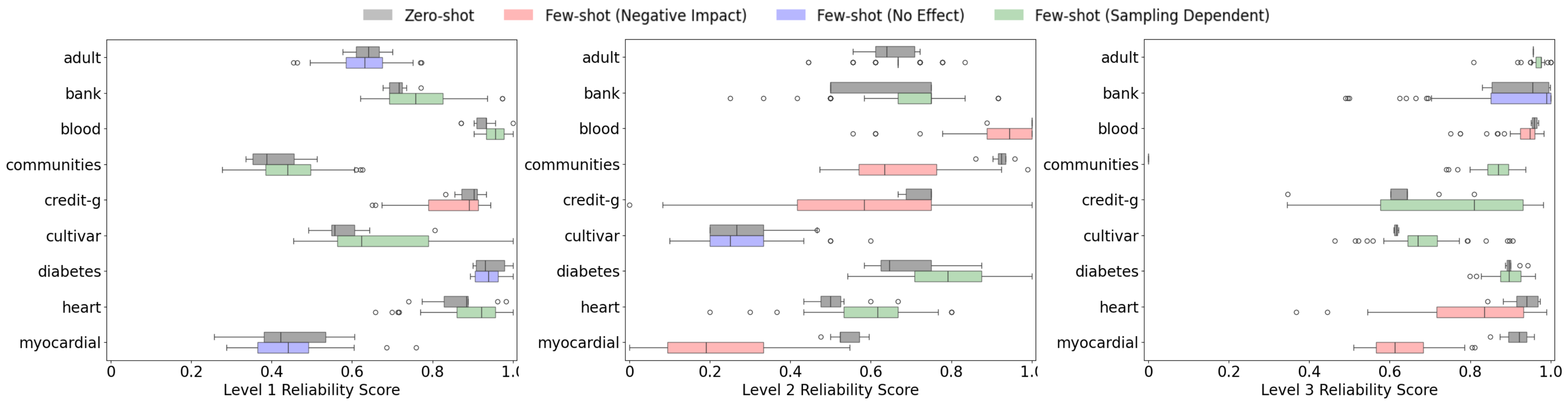}
  \caption {\chgd{Effects of varying the number of examples on reliability scores at each level (with Communities).}}
  \label {fig:experiments_communities1}
  \vspace{-0.5cm}
\end {figure*}

%% file: Tables/Table_4.tex
\begin{table*}
\centering
\caption{Few-shot classification performance evaluation results on the Communities dataset.}
\label{tab:performance_communities}
\resizebox{\textwidth}{!}{
\begin {tabular}{c|c|ccc|cccc}
\toprule
\ {Data}           & Shot                   & LogReg           & RandomForest          & XGBoost     & FeatLLM          & Ours (Top 3)       & Ours (Top 5) & Improvement (\%) \\
\midrule
\multirow{3}{*}{Communities}      
                            & 4                      & $47.95\pm2.07$ & $54.94\pm7.06$ & $50.0\pm0.0$ & $\underline{73.82\pm2.93}$ & $72.26\pm8.69$ & $\bold{74.77\pm5.88}$ & $\textcolor {blue}{\blacktriangledown} 0.99$\\ 
                            & 8                      & $55.81\pm10.46$ & $64.09\pm4.39$ & $68.82\pm3.26$ & $\underline{74.88\pm3.73}$ & $74.64\pm5.46$ & $\bold {76.5\pm3.25}$ & $\textcolor{red}{\blacktriangle} 1.62$\\ 
                            & 16                      & $59.23\pm13.75$ & $68.82\pm0.69$ & $66.03\pm0.96$ & $73.88\pm2.87$ & $\underline{73.91\pm3.79}$ & $\bold{74.77\pm2.64}$ & $\textcolor{red}{\blacktriangle} 0.89$\\
\bottomrule
\end{tabular}
}
\end{table*}

%% file: Figures/Figure_A7.tex
\begin {figure*}[t!]
  \centering
   \begin{minipage}{0.49\textwidth}
  \includegraphics [width=\columnwidth]{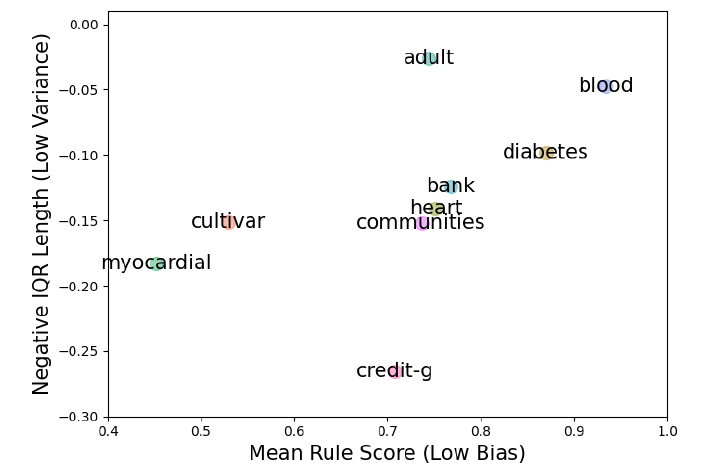}
  \caption {\chgd{Variance and bias of average reliability score  (with Communities).}}
  \label {fig:experiments_communities2}
  \end {minipage}
  \hfill
  \begin{minipage}{0.49\textwidth}
  \includegraphics [width=\columnwidth]{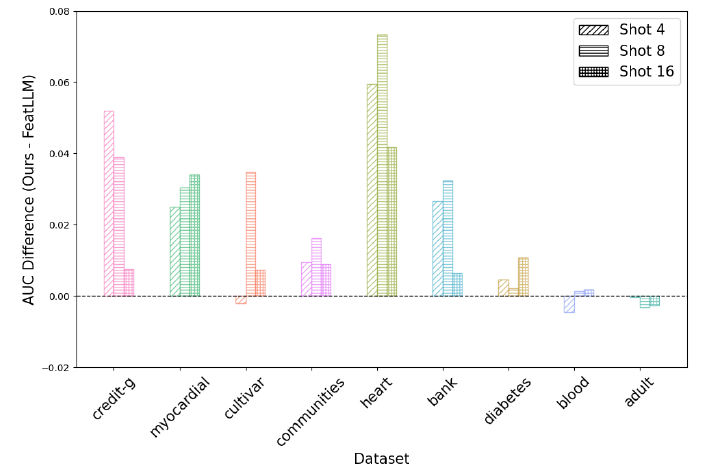}
  \caption {\chgd{Performance improvement of ours over
  \label {fig:experiments_communities3} FeatLLM with different shots (with Communities).}}
  \end{minipage}
  \vspace{-0.5cm}
\end {figure*}

%% file: Figures/Figure_A3.tex
\begin {figure*}[hbt!]
  \includegraphics [width=\textwidth]{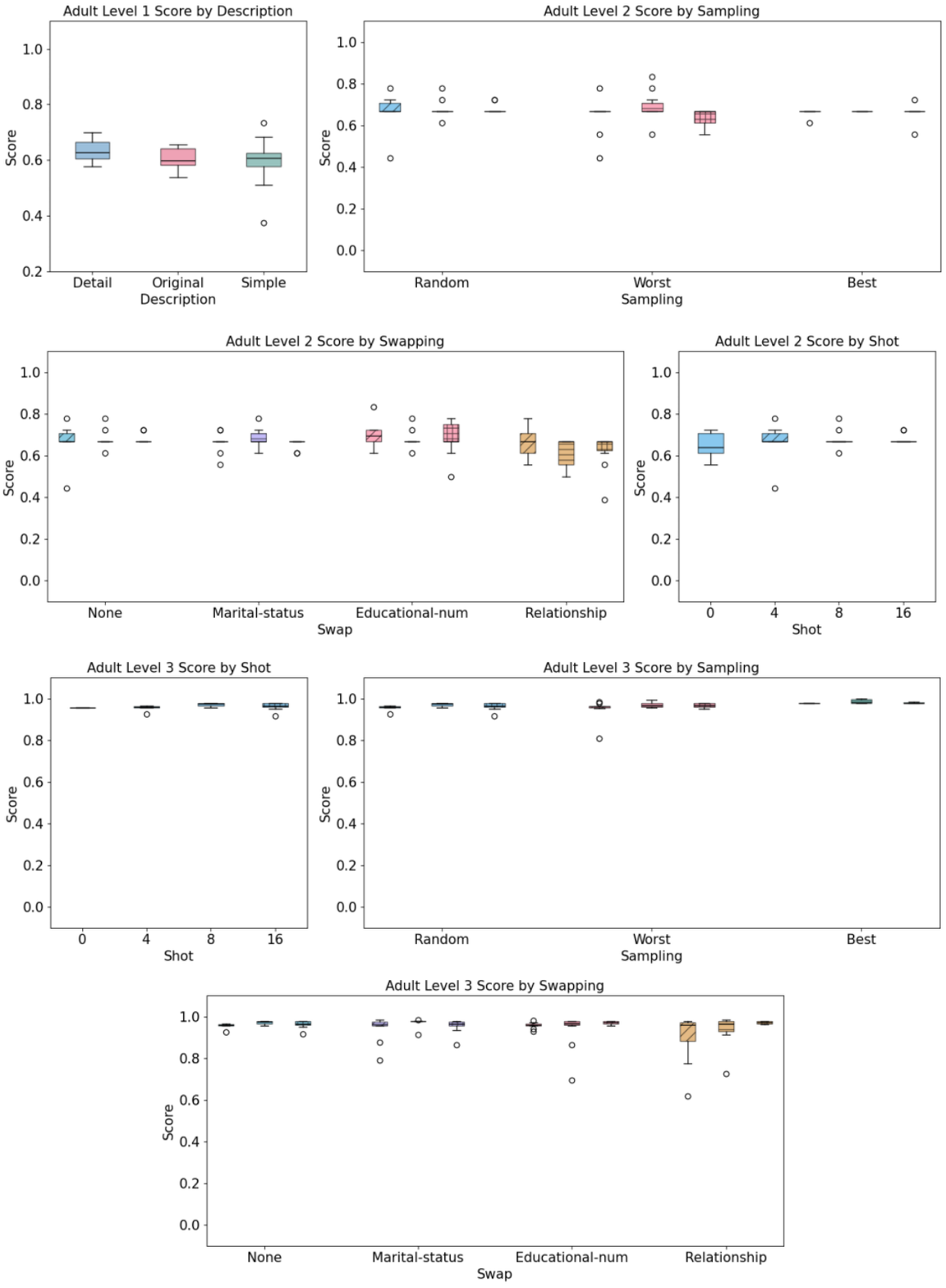}
  \caption {Full results of reliability diagnosis on Adult.}
  \label {fig:adult}
\end {figure*}

\begin {figure*}[hbt!]
  \includegraphics [width=\textwidth]{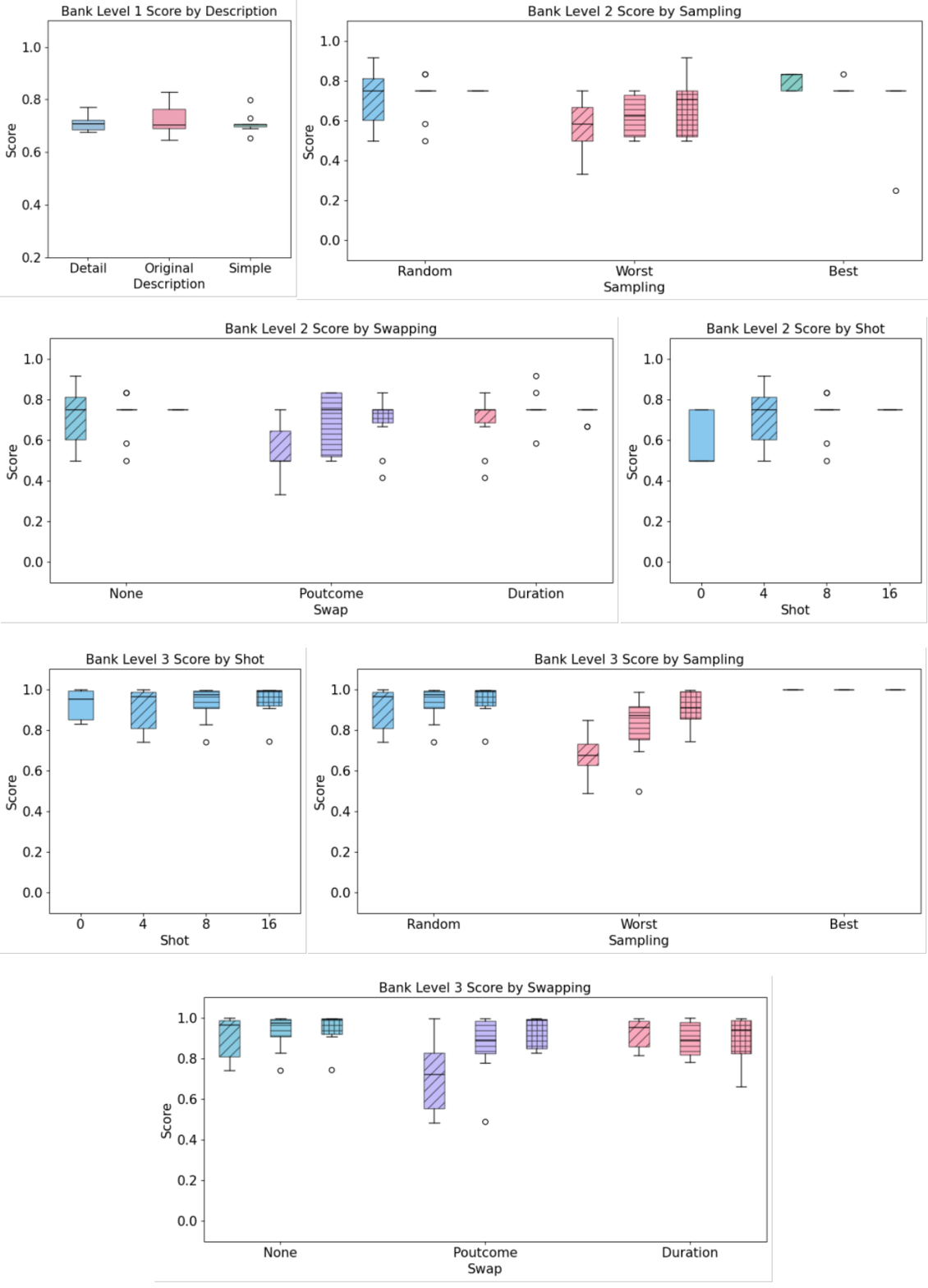}
  \caption {Full results of reliability diagnosis on Bank.}
  \label {fig:bank}
\end {figure*}

\begin {figure*}[hbt!]
  \includegraphics [width=\textwidth]{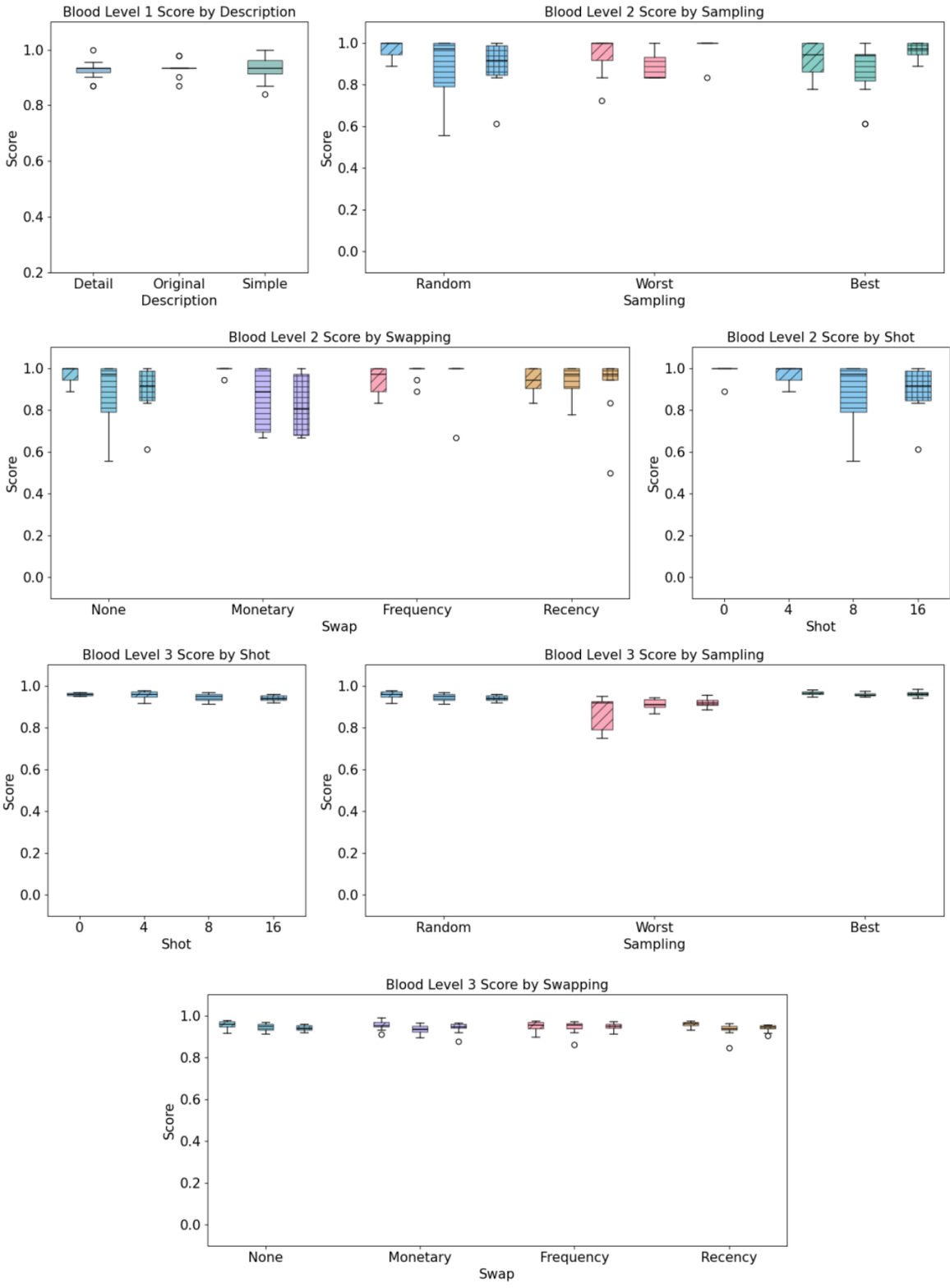}
  \caption {Full results of reliability diagnosis on Blood.}
  \label {fig:blood}
\end {figure*}

\begin {figure*}[hbt!]
  \includegraphics [width=\textwidth]{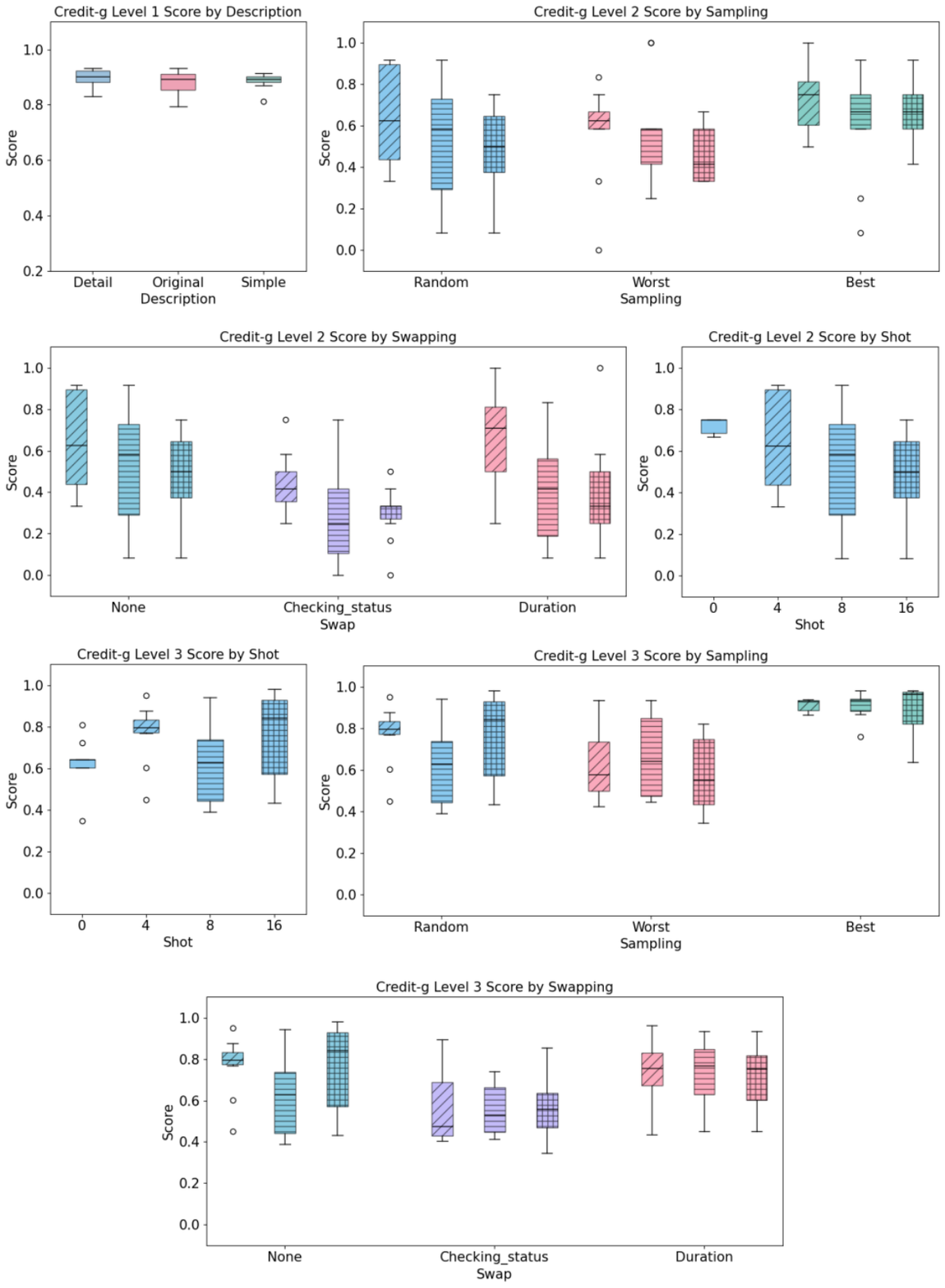}
  \caption {Full results of reliability diagnosis on Credit-g.}
  \label {fig:credit-g}
\end {figure*}

\begin {figure*}[hbt!]
  \includegraphics [width=\textwidth]{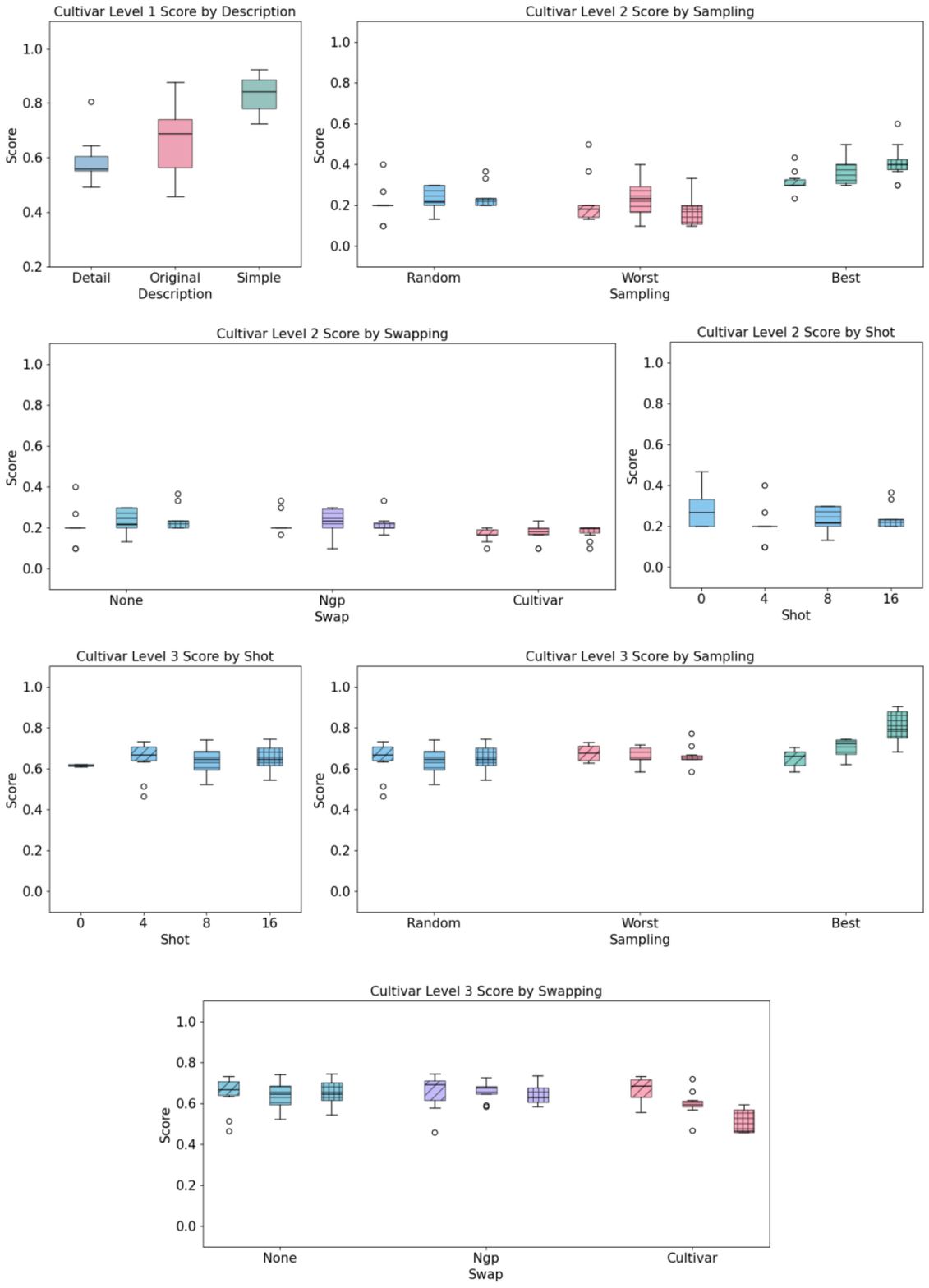}
  \caption {Full results of reliability diagnosis on Cultivar.}
  \label {fig:cultivar}
\end {figure*}

\begin {figure*}[hbt!]
  \includegraphics [width=\textwidth]{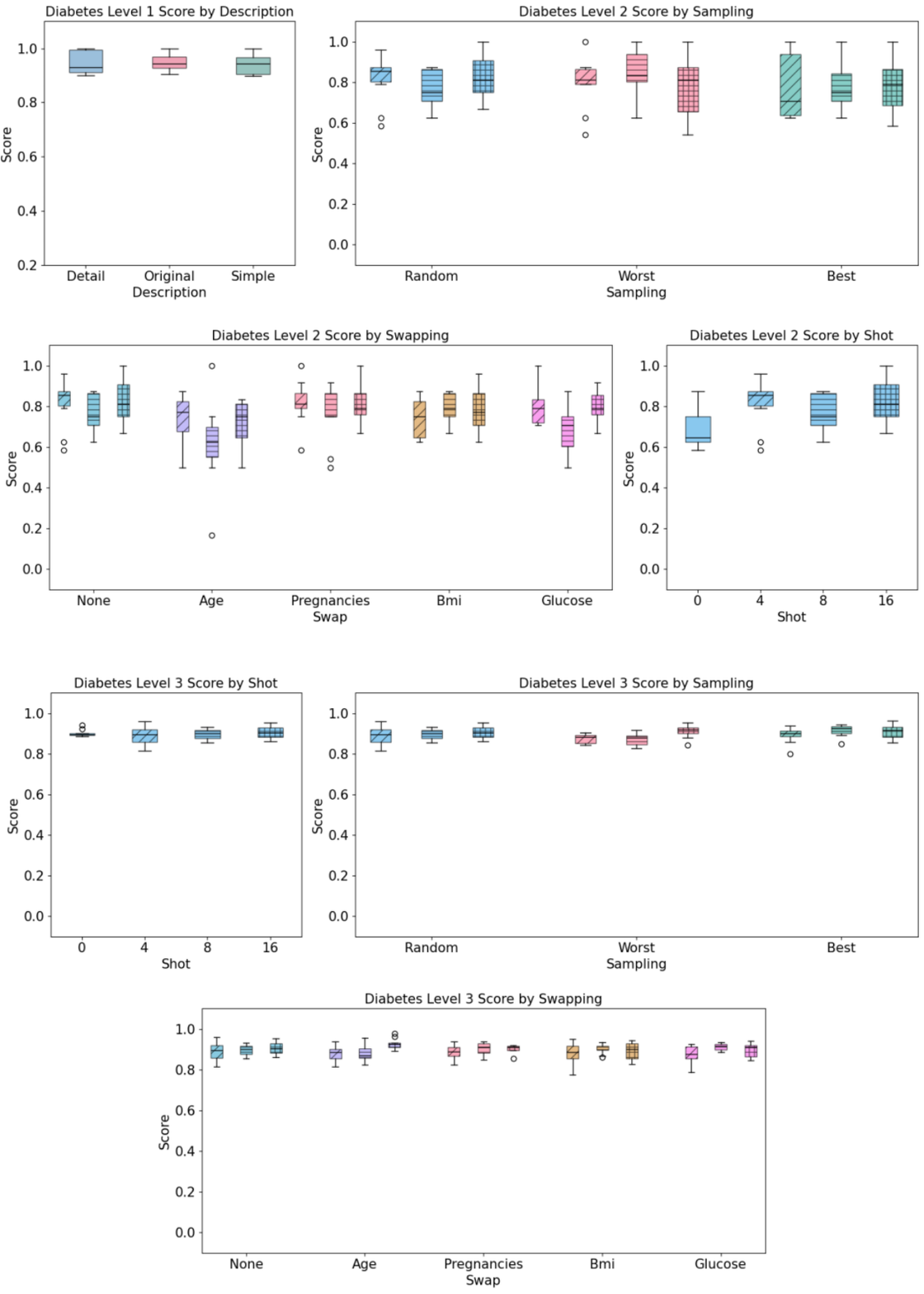}
  \caption {Full results of reliability diagnosis on Diabetes.}
  \label {fig:diabetes}
\end {figure*}

\begin {figure*}[hbt!]
  \includegraphics [width=\textwidth]{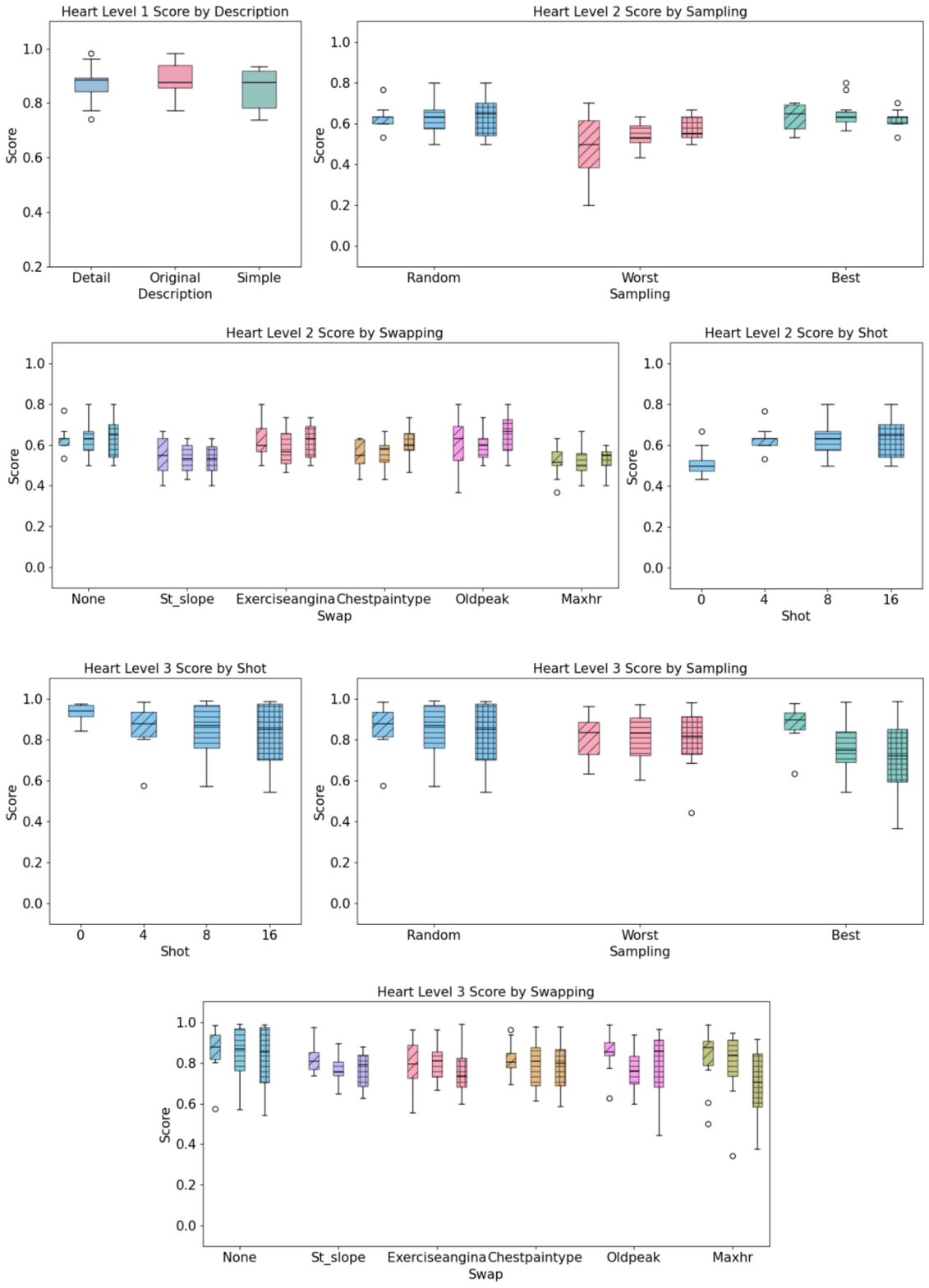}
  \caption {Full results of reliability diagnosis on Heart.}
  \label {fig:heart}
\end {figure*}

\begin {figure*}[hbt!]
  \includegraphics [width=\textwidth]{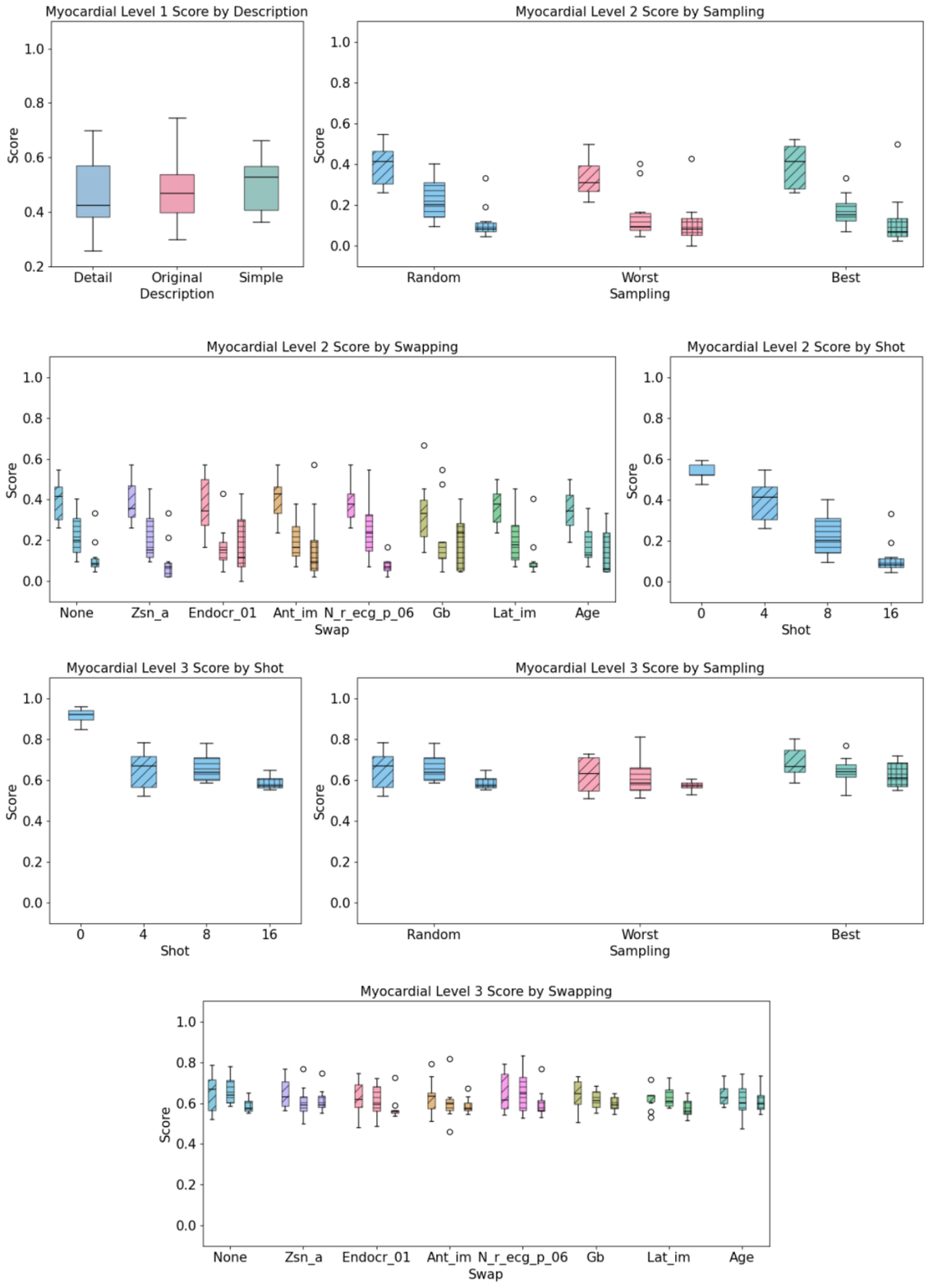}
  \caption {Full results of reliability diagnosis on Myocardial.}
  \label {fig:myocardial}
\end {figure*}

%% file: Figures/Prompt_A1.tex
\begin{figure*}[tb]

\begin{tcolorbox}[colback=black!5,colframe=black!40]
You are an expert. Given the task description and the list of features, you are ranking them according to their importances to solve the task. The ranking should be in descending order, starting with the most important feature.\\

Task: Does the coronary angiography of this patient show a heart disease? Yes or no?\\

Features:
\begin{itemize}
\item Age: age of the patient (numerical variable)
\item Sex: sex of the patient (categorical variable with categories [M, F])
\item ChestPainType: chest pain type (categorical variable with categories [ATA, NAP, ASY, TA])
\item RestingBP: resting blood pressure [mm Hg] (numerical variable)
\item Cholesterol: serum cholesterol [mm/dl] (numerical variable)
\item FastingBS: fasting blood sugar [1: if FastingBS > 120 mg/dl, 0: otherwise] (numerical variable)
\item RestingECG: resting electrocardiogram results (categorical variable with categories [Normal, ST, LVH])
\item MaxHR: maximum heart rate achieved (numerical variable)
\item ExerciseAngina: exercise-induced angina (categorical variable with categories [N, Y])
\item Oldpeak: oldpeak = ST [Numeric value measured in depression] (numerical variable)
\item ST\_Slope: the slope of the peak exercise ST segment (categorical variable with categories [Up, Flat, Down])
\end{itemize}

Your response should be a numbered list with each item on a new line.\\

Format for Response:
\begin{verbatim}
Rank:
FeatA
FeatB....

\end{verbatim}

Answer:
\end{tcolorbox}
\caption{Example prompt for reliability diagnosis level 1 for Heart dataset.}
\label{fig:prompt1}
\end{figure*}

%% file: Figures/Prompt_A2.tex
\begin{figure*}[tb]
\begin{tcolorbox}[colback=black!5,colframe=black!40, width=\textwidth]
You are an expert. Given the task description and the list of features and data examples, analyze the causal relationship or tendency between each feature and class based on general knowledge and common sense within a short sentence.\\

Task: Does the coronary angiography of this patient show a heart disease? Yes or no?\\

Features:
\begin{itemize}[nosep]
\item ChestPainType: chest pain type (categorical variable with categories [ATA, NAP, ASY, TA])
\item MaxHR: maximum heart rate achieved (numerical variable)
\item Oldpeak: oldpeak = ST [Numeric value measured in depression] (numerical variable)
\item ST\_Slope: the slope of the peak exercise ST segment (categorical variable with categories [Up, Flat, Down])
\item ExerciseAngina: exercise-induced angina (categorical variable with categories [N, Y])\\
\end{itemize}

Examples:
\begin{verbatim}
Age is 63. Sex is M. ChestPainType is NAP. RestingBP is 130. Cholesterol is 0.
FastingBS is 1. RestingECG is ST. MaxHR is 160. ExerciseAngina is Y. 
Oldpeak is 3.0. ST_Slope is Flat.
Answer: no
Age is 39. Sex is M. ChestPainType is ATA. RestingBP is 120. Cholesterol is 204.
FastingBS is 0. RestingECG is Normal. MaxHR is 145. ExerciseAngina is N. 
Oldpeak is 0.0. ST_Slope is Up.
Answer: no
Age is 55. Sex is M. ChestPainType is ASY. RestingBP is 160. Cholesterol is 289.
FastingBS is 0. RestingECG is LVH. MaxHR is 145. ExerciseAngina is N. 
Oldpeak is 0.8. ST_Slope is Flat.
Answer: yes
Age is 58. Sex is M. ChestPainType is NAP. RestingBP is 160. Cholesterol is 211. 
FastingBS is 1. RestingECG is ST. MaxHR is 92. ExerciseAngina is N. 
Oldpeak is 0.0. ST_Slope is Flat.
Answer: yes
\end{verbatim}

Format for Response:
\begin{verbatim}
Causal Relationship for class "no":
ChestPainType in ATA has a [positive/negative] correlation with class no
MaxHR has a [positive/negative] correlation with class no
Oldpeak has a [positive/negative] correlation with class no
ST_Slope in Up has a [positive/negative] correlation with class no
ExerciseAngina in N has a [positive/negative] correlation with class no

Causal Relationship for class "yes":
ChestPainType in ASY has a [positive/negative] correlation with class yes
MaxHR has a [positive/negative] correlation with class yes
Oldpeak has a [positive/negative] correlation with class yes
ST_Slope in Flat has a [positive/negative] correlation with class yes
ExerciseAngina in Y has a [positive/negative] correlation with class yes

\end{verbatim}

Answer:
\end{tcolorbox}
\caption{Example prompt for reliability diagnosis level 2 for Heart dataset. }
\label{fig:prompt2}

\end{figure*}

%% file: Figures/Prompt_A3.tex
\begin{figure*}[tb]
\begin{tcolorbox}[colback=black!5,colframe=black!40]
You are an expert. Given the task description and the list of features and data examples, you are filling in the feature conditions for each class to solve the task.\\

Task: Does the coronary angiography of this patient show a heart disease? Yes or no?\\

Features:
\begin{itemize}[nosep]
\item ChestPainType: chest pain type (categorical variable with categories [ATA, NAP, ASY, TA])
\item MaxHR: maximum heart rate achieved (numerical variable)
\item Oldpeak: oldpeak = ST [Numeric value measured in depression] (numerical variable)
\item ST\_Slope: the slope of the peak exercise ST segment (categorical variable with categories [Up, Flat, Down])
\item ExerciseAngina: exercise-induced angina (categorical variable with categories [N, Y])\\
\end{itemize}

Examples:
\begin{verbatim}
Age is 39. Sex is M. ChestPainType is ATA. RestingBP is 120. Cholesterol is 204. 
FastingBS is 0. RestingECG is Normal. MaxHR is 145. ExerciseAngina is N. 
Oldpeak is 0.0. ST_Slope is Up. Answer: no
Age is 63. Sex is M. ChestPainType is NAP. RestingBP is 130. Cholesterol is 0. 
FastingBS is 1. RestingECG is ST. MaxHR is 160. ExerciseAngina is Y. 
Oldpeak is 3.0. ST_Slope is Flat. Answer: no
Age is 55. Sex is M. ChestPainType is ASY. RestingBP is 160. Cholesterol is 289. 
FastingBS is 0. RestingECG is LVH. MaxHR is 145. ExerciseAngina is N. 
Oldpeak is 0.8. ST_Slope is Flat. Answer: yes
Age is 58. Sex is M. ChestPainType is NAP. RestingBP is 160. Cholesterol is 211. 
FastingBS is 1. RestingECG is ST. MaxHR is 92. ExerciseAngina is N. 
Oldpeak is 0.0. ST_Slope is Flat. Answer: yes
\end{verbatim}

Format for Response:
\begin{verbatim}
Condition for class "no":
ChestPainType is in [Value]
MaxHR is greater than [Value]
Oldpeak is less than [Value]
ST_Slope is in [Value]
ExerciseAngina is in [Value]

Condition for class "yes":
ChestPainType is in [Value]
MaxHR is less than [Value]
Oldpeak is greater than [Value]
ST_Slope is in [Value]
ExerciseAngina is in [Value]

\end{verbatim}
Format for \texttt{[Value]}:
\begin{itemize}[nosep]
\item For the categorical variable only: \texttt{[List of Categories]}
\item For the numerical variable only: \texttt{[Value]}\\
\end{itemize}

Answer:
\end{tcolorbox}
\caption{Example prompt for reliability diagnosis level 3 for Heart dataset. }
\label{fig:prompt3}

\end{figure*}

%% file: Figures/Prompt_A4.tex
\begin{figure*}[tb]
\begin{tcolorbox}[colback=black!5,colframe=black!40]

You are an expert. Given the task description and the list of features and data examples, you are extracting conditions for each answer class to solve the task.\\

Task: \texttt{[TASK]}\\

Features: \texttt{[FEATURES]}\\

Examples: \texttt{[EXAMPLES]}\\

Let's first understand the problem and solve the problem step by step.

Step 1. Analyze the causal relationship or tendency between each feature and task description based on general knowledge and common sense within a short sentence. \\
Step 2. Based on the above examples and Step 1 results, infer 10 different conditions per answer, following the format below. The condition should make sense, well match examples, and must match the format for \texttt{[Condition]} according to value type. \\

Format for Response:\\
10 different conditions for class “no”:\\
- \texttt{[Condition]}\\
... 

10 different conditions for class “yes”:\\
- \texttt{[Condition]}\\
...\\

Format for \texttt{[Condition]}:
\begin{itemize}[nosep, leftmargin=10pt]
  \item For the categorical variable only:\\
    \texttt{[Feature] is in [list of Categories]}
  \item For the numerical variable only:\\
    \texttt{[Feature] (> or >= or < or <=) [Value]}\\
    \texttt{[Feature] is within range of [Value start, Value end]} \\
\end{itemize}

Answer: Step 1.

\end{tcolorbox}
\caption{Prompt for default feature engineering in FeatLLM. }
\label{fig:prompt4}
\end{figure*}